%% file: MMLA.tex
\documentclass{article}
\PassOptionsToPackage{numbers, compress}{natbib}
\usepackage[preprint]{neurips_2025}
\usepackage{enumitem}
\usepackage{microtype}
\usepackage{multirow}
\usepackage{makecell}
\usepackage[utf8]{inputenc} % allow utf-8 input
\usepackage[T1]{fontenc}    % use 8-bit T1 fonts
\usepackage{hyperref}       % hyperlinks
\usepackage{url}            % simple URL typesetting
\usepackage{booktabs}       % professional-quality tables
\usepackage{amsfonts}       % blackboard math symbols
\usepackage{amsmath}
\usepackage{nicefrac}       % compact symbols for 1/2, etc.
\usepackage{microtype}      % microtypography
\usepackage{xcolor}         % colors
\usepackage{graphicx} % Required for inserting images
\usepackage{subcaption} % 关键包：支持子图
\usepackage{booktabs}   % 用于三线表格式
\usepackage{array}      % 用于列格式定义
\usepackage{multirow}   % 用于多行合并
\usepackage{wrapfig}
\usepackage{inconsolata}
\usepackage{times}
\usepackage{latexsym}
\usepackage{alltt}   % 用于展示prompt

\title{Can Large Language Models Help Multimodal Language Analysis? MMLA: A Comprehensive Benchmark}

\author{
Hanlei Zhang$^1$\footnotemark[1]\quad
Zhuohang Li$^1$\footnotemark[1]\quad
Yeshuang Zhu$^{2}$\quad
Hua Xu$^{1}$\footnotemark[2]\\  
\textbf{Peiwu Wang}$^{1}$\quad
\textbf{Haige Zhu}$^{3}$\quad
\textbf{Jie Zhou}$^{2}$\quad
\textbf{Jinchao Zhang}$^{2}$\quad
 \\
$^1$Department of Computer Science and Technology, Tsinghua University \\
$^2$Pattern Recognition Center, WeChat AI, Tencent Inc, China \quad
$^3$Kennesaw State University 
}

\begin{document}
\renewcommand{\thefootnote}{\fnsymbol{footnote}} % 设置脚注为符号样式
\maketitle
\footnotetext[1]{Equal contribution.} % 添加脚注内容
\footnotetext[2]{Corresponding author.} % 添加脚注内容

\setcounter{footnote}{0} % 重置脚注计数器
\renewcommand{\thefootnote}{\arabic{footnote}} % 将脚注样式恢复为数字

\maketitle
\input{sec/0_abs}    
\input{sec/1_intro}
\input{sec/2_related_work}

\input{sec/3_benchmark}

\input{sec/4_experiments}

\input{sec/5_discussions}
\input{sec/6_conclusions}
\bibliographystyle{plainnat}\small  % 或者 neurips_2024 中自带的样式
\bibliography{MMLA}

\input{sec/supplementary}
\end{document}

%% file: sec/0_abs.tex
\begin{abstract}
Multimodal language analysis is a rapidly evolving field that leverages multiple modalities to enhance the understanding of high-level semantics underlying human conversational utterances. Despite its significance, little research has investigated the capability of multimodal large language models (MLLMs) to comprehend cognitive-level semantics. In this paper, we introduce MMLA, a comprehensive benchmark specifically designed to address this gap. MMLA comprises over 61K multimodal utterances drawn from both staged and real-world scenarios, covering six core dimensions of multimodal semantics: intent, emotion, dialogue act, sentiment, speaking style, and communication behavior. We evaluate eight mainstream branches of LLMs and MLLMs using three methods: zero-shot inference, supervised fine-tuning, and instruction tuning. Extensive experiments reveal that even fine-tuned models achieve only about 60\%$\sim$70\% accuracy, underscoring the limitations of current MLLMs in understanding complex human language. We believe that MMLA will serve as a solid foundation for exploring the potential of large language models in multimodal language analysis and provide valuable resources to advance this field. The datasets and code are open-sourced at https://github.com/thuiar/MMLA.

\end{abstract}

%% file: sec/1_intro.tex
\section{Introduction}
\begin{figure*}[htbp]
  \centering
  \includegraphics[width=\textwidth]{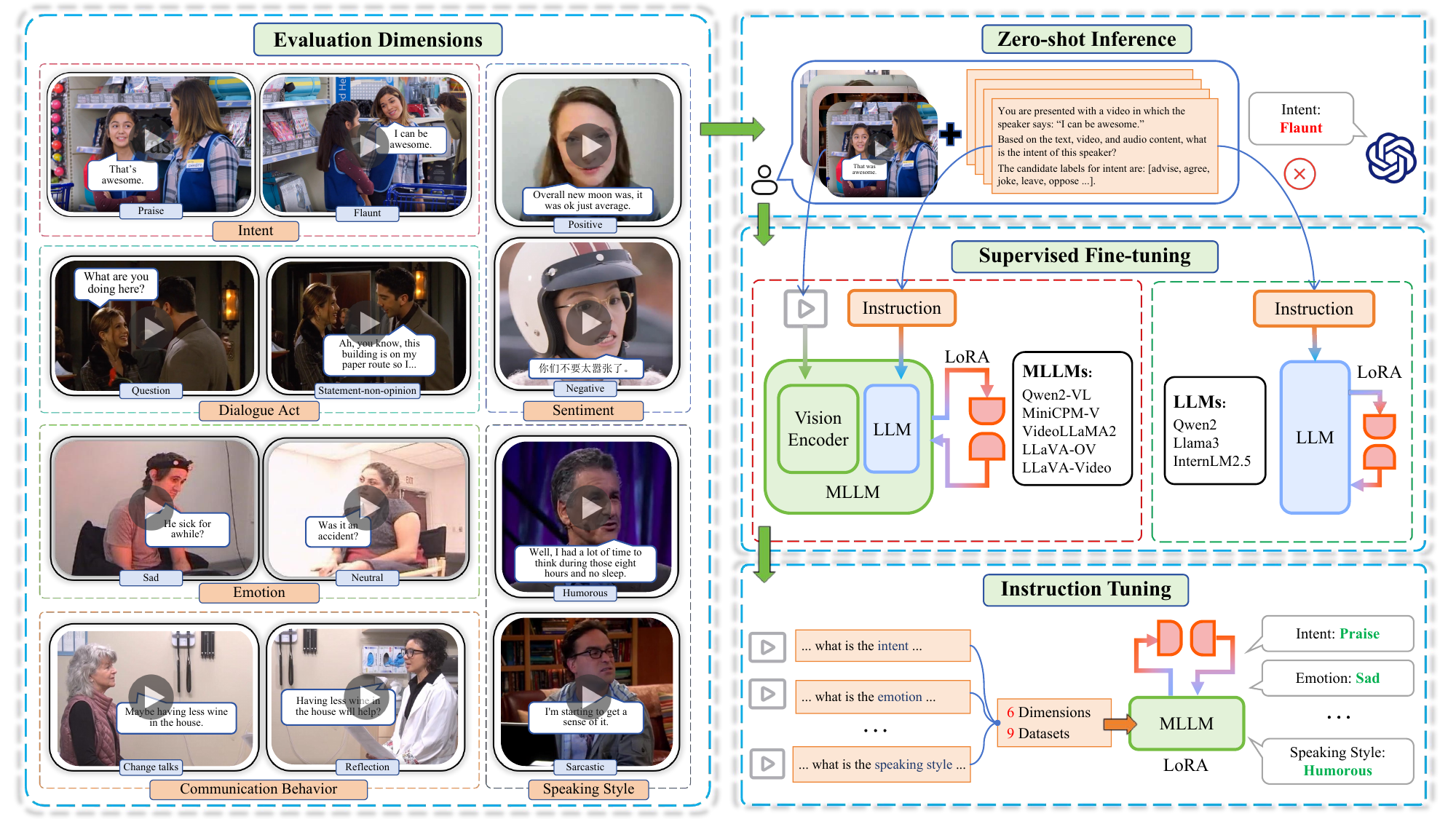}
\caption{Overview of the MMLA benchmark. The left side shows examples from six evaluation dimensions and nine datasets. The right side displays three methods for evaluating both LLMs and MLLMs: (1) zero-shot inference (top right), which generates predictions from task-specific prompts; (2) supervised fine-tuning (middle right), which trains on each supervised task; and (3) instruction tuning (bottom right), which trains on multiple tasks simultaneously. Both (2) and (3) utilize LoRA to efficiently adapt foundation models.}
  \label{fig:MMLA}
  \vspace{-3.5mm} % 根据需要调整负空间
\end{figure*}

Multimodal language analysis has emerged as a prominent research area~\cite{das2023multimodal}, utilizing various modalities to decode cognitive-level semantics in human utterances (e.g., emotion, intent). This analysis is crucial for understanding psychological and behavioral motivations, and it has broad applications in virtual assistants~\cite{wu-etal-2023-simmc}, recommender systems~\cite{cui2019user}, and social behavior analysis~\cite{mittal2024towards}.

This field has attracted significant attention, with early works focusing on annotating sentiment intensity from social media videos~\cite{zadeh2016mosi,zadeh2018multimodal} and conversations from various TV shows or movies~\cite{yu-etal-2020-ch,liu2022make}. Additionally, researchers have provided emotion categories for TV shows~\cite{poria-etal-2019-meld} based on Ekman's six universal emotions~\cite{ekman1980facial}. Building on these resources, numerous methods have been developed to learn complementary information and alleviate the challenges posed by the heterogeneous nature of different modalities~\cite{tsai2019multimodal,hazarika2020misa,hu-etal-2022-unimse,zhang-etal-2024-amanda}. In addition to sentiment and emotion, researchers have investigated other linguistic properties such as sarcasm~\cite{castro2019towards,zhao2023multi} and humor~\cite{hasan2019ur,10707160}, with multimodal fusion methods specifically designed for binary classification tasks~\cite{hasan2021humor,pramanick2022multimodal}. More recently, studies have focused on analyzing coarse-grained and fine-grained intents using new datasets and taxonomies~\cite{saha2020towards,mintrec,zhang2024mintrec2.0}, although this area is still in its early stages~\cite{Sun_2024_CVPR,zhu2024inmu}.

Despite these advances, existing methods predominantly rely on fusion techniques built on lightweight neural networks~\cite{rahman-etal-2020-integrating,yu2021learning,chen-etal-2024-d2r}, which show limited performance on more complicated reasoning tasks~\cite{yin2023survey}. The advent of MLLMs~\cite{li2023blip,liu2023visual,zhuminigpt,NEURIPS2024_dc06d4d2} reveals the huge potential for emergent cross-modal reasoning capabilities through scalable model parameters~\cite{yin2023survey}. However, existing MLLM benchmarks mainly focus on low-level perceptual semantics, such as scene and procedure understanding~\cite{li2024seed}, instance location~\cite{liu2025mmbench}, and elementary cognitive-level tasks like video content analysis~\cite{fu2024video} and commonsense reasoning~\cite{yin2023survey}. These benchmarks fail to address high-level semantics in conversations. Other benchmarks in this field include only a few semantic dimensions, such as emotion and intent~\cite{yang2024emollm,liu2024emotion}, or are incapable of evaluating LLMs~\cite{liang2021multibench}.

To address these challenges, we propose MMLA, the first comprehensive benchmark for multimodal language analysis, aimed at evaluating foundation models. Figure~\ref{fig:MMLA} provides an overview of MMLA. In this benchmark, we introduce six representative semantic dimensions for evaluation: \textit{intent}, \textit{emotion}, \textit{dialogue act}, \textit{sentiment}, \textit{speaking style}, and \textit{communication behavior}. These dimensions cover the most important cognitive-level semantic aspects of multimodal conversational interactions. We then collect nine publicly available multimodal language datasets, totaling over 61K multimodal utterances across more than 76 hours of video, with each utterance containing text, video, and audio modalities. These datasets span various sources, including both staged scenarios (e.g., TV series, films, TED talks) and real-world settings (e.g., spontaneous social media videos and motivational interviews). Next, we evaluate state-of-the-art (SOTA) LLMs and MLLMs on MMLA. In particular, five branches of MLLMs are employed to leverage both language and video modalities. Additionally, three branches of LLMs that process only text are used for comparison with MLLMs, to assess the effect of non-verbal modalities. The sizes of these models range from 0.5B to 72B parameters. We apply zero-shot inference, supervised fine-tuning, and instruction tuning as evaluation methods.

Extensive experiments demonstrate that existing MLLMs show limited performance in understanding high-level cognitive semantics. Supervised fine-tuning can significantly enhance the multimodal capabilities and achieve new SOTA performance on most tasks. In particular, smaller-scale models show great potential with performance comparable to that of larger-scale models. Through instruction tuning, foundation models can successfully handle multiple tasks with a unified model, achieving performance comparable to supervised fine-tuning. However, even after tuning, these models still exhibit a significant limitation on these tasks, with average accuracy scores below 70\%.

Our contributions are summarized as follows: 
(1) We propose MMLA, a large-scale multimodal language analysis benchmark containing 61K multimodal utterances drawn from over 76 hours of video. MMLA spans six core dimensions that are crucial for understanding high-level cognitive semantics. (2) To the best of our knowledge, MMLA is the first to comprehensively assess the capabilities of foundation models in multimodal language analysis, by evaluating nine mainstream models across three strategies. (3) Extensive experiments reveal new insights into foundation models for multimodal language analysis. MMLA also pushes the limits of existing MLLMs, providing a solid basis and promising new directions for further research. The code is publicly available, and the data are released under their respective licenses (see Appendix \ref{appendix:license} for details).

%% file: sec/2_related_work.tex
\section{Related Works} 
\textbf{Multimodal Language Datasets}. With the boom in multimodal language analysis, many significant tasks have emerged alongside the development of benchmark datasets. For example, early research focused on multimodal sentiment analysis and emotion recognition, and there are numerous datasets designed to analyze multilingual opinion sentiment~\cite{perez-rosas-etal-2013-utterance,zadeh2016mosi,yu-etal-2020-ch,liu2022make}.~\citet{zadeh2018multimodal} constructed the first large-scale dataset in this field and additionally annotated emotion labels following Ekman’s taxonomies~\cite{ekman1980facial}. However, these datasets only involve individual opinions and lack conversational interactions among multiple speakers.~\citet{busso2008iemocap} introduced a dataset that records conversations between two speakers and annotates each utterance with emotion labels from nine categories in multimodal contexts. Nonetheless, dyadic sessions pose limitations when dealing with real-world multi-party scenarios. To address this,~\citet{poria-etal-2019-meld} provided sentiment and emotion labels for conversations taken from a TV series involving multiple speakers. These abundant resources have led to extensive research on designing effective multimodal fusion methods, including tensor operation-based~\cite{zadeh-etal-2017-tensor,liu-etal-2018-efficient-low,zadeh2018memory} and transformer-based approaches~\cite{tsai2019multimodal,rahman-etal-2020-integrating,hazarika2020misa,han2021improving,hu-etal-2022-unimse,zhang2024unsupervised}.

Beyond the relatively shallow semantics of sentiment or emotion, researchers have begun to explore more diverse and complex intent semantics in utterances, resulting in substantial new resources. Early work in this field analyzed authors' intents on social media platforms. For instance,~\citet{kruk-etal-2019-integrating} proposed a taxonomy of eight intents based on rhetorical classes, and \citet{zhang2021multimet} introduced four intent classes related to metaphor. However, these intents differ from those found in conversational scenarios. \citet{saha2020towards} annotated 12 dialogue acts drawn from the Switchboard\cite{godfrey1992switchboard} tag set for two multimodal emotion recognition datasets~\cite{busso2008iemocap,poria-etal-2019-meld}. Nevertheless, these dialogue acts are coarse-grained communicative intents that are not directly applicable to real-world applications~\cite{mintrec}. To address this issue, \citet{mintrec} proposed the first hierarchical intent taxonomy specifically designed for multimodal contexts and introduced the first multimodal conversational intent recognition dataset. \citet{zhang2024mintrec2.0} subsequently extended this dataset into a larger-scale version that accommodates multi-party interactions and includes out-of-scope utterances, reflecting real-world conditions. In addition, some research has focused on individual speaking styles, such as humor~\cite{hasan2019ur} and sarcasm~\cite{castro2019towards}, which are driven by particular human intents, such as joking or mocking. Recently,~\citet{Anno-MI} investigated more complex communication behaviors between clients and therapists through motivational interviewing in counseling scenarios.

\textbf{Benchmarks}. There are also multimodal benchmarks related to this work. For example, MultiBench~\cite{liang2021multibench} constructs a large-scale multimodal learning benchmark spanning various areas, such as healthcare and robotics. Nevertheless, it only covers dimensions related to affective computing and evaluates traditional multimodal machine learning methods without incorporating powerful MLLMs. To investigate the capability of MLLMs, numerous benchmarks have been proposed in recent years. However, most benchmarks focus on perceptual-level or elementary cognitive-level tasks such as visual recognition~\cite{li2024seed}, optical character recognition~\cite{fu2024video}, multimodal question answering~\cite{ning2023video}, video content analysis~\cite{ataallah2024infinibench,fang2024mmbench}, scientific calculation~\cite{liu2025mmbench}, and visual reasoning~\cite{liu2023visual}. While previous benchmarks cover diverse domains and tasks, none specifically target large-scale multimodal language analysis. MMLA is the first benchmark designed to advance this field in foundation models. 

\textbf{Multimodal Large Language Models}. Multimodal large language models have emerged as a new paradigm in multimodal learning due to their superior scalability and cross-modal reasoning capabilities. For example, VideoLLaMA2~\cite{cheng2024videollama} introduces a Spatio-Temporal Convolution (STC) connector that excels in capturing spatiotemporal dynamics for audio-visual tasks. LLaVA-OneVision~\cite{li2024llava} pioneers cross-modal transfer learning, excelling in zero-shot video understanding despite being trained only on image datasets. LLaVA-Video~\cite{zhang2024video} introduces a new video representation technique that allows maximizing the sampling of video frames, and a high-quality dataset is constructed to promote its instruction-following capability. Qwen2-VL~\cite{wang2024qwen2} leads in vision-language understanding and generation, performing well in both zero-shot and few-shot settings. MiniCPM-V~\cite{yao2024minicpm} innovates in model compression to enable efficient mobile deployment without compromising performance. Although current MLLMs perform well on various tasks, no benchmark evaluates their ability to handle complex multimodal language analysis.

%% file: sec/3_benchmark.tex
\section{MMLA Benchmark}

\begin{table*}
\centering
\scalebox{0.88}{
\scriptsize
\setlength{\tabcolsep}{3pt}
\begin{tabular}{cccccccccccc}
\toprule
\textbf{Dimensions} & \textbf{Datasets} & \textbf{\#C} & \textbf{\#U} & \textbf{\#Train} & \textbf{\#Val} & \textbf{\#Test} & \textbf{\makecell{Video \\ Hours}} & \textbf{Source} & \makecell{\textbf{\#Video Length} \\ avg. / max.} & \makecell{\textbf{\#Text Length} \\ avg. / max.} & \textbf{Language} \\ 
\midrule
\multirow{2}{*}{Intent} 
& MIntRec & 20 & 2,224 & 1,334 & 445 & 445  & 1.5 & TV series & 2.4 / 9.6 & 7.6 / 27.0 & \multirow{2}{*}{English}  \\
& MIntRec2.0 & 30 & 9,304 & 6,165 & 1,106 & 2,033  & 7.5 & TV series & 2.9 / 19.9 & 8.5 / 46.0 &  \\ 
\addlinespace[0.05cm]
\cline{1-12}
\addlinespace[0.05cm]
\multirow{2}{*}{\makecell{Dialogue \\ Act}}
& MELD & 12 & 9,989 & 6,992 & 999 & 1,998 & 8.8  & TV series & 3.2 / 41.1 & 8.6 / 72.0 & \multirow{2}{*}{English} \\
& IEMOCAP & 12 & 9,416 & 6,590 & 942 & 1,884 & 11.7 & Improvised scripts & 4.5 / 34.2 & 12.4 / 106.0 &  \\ 
\addlinespace[0.05cm]
\cline{1-12}
\addlinespace[0.05cm]
\multirow{2}{*}{Emotion} 
& MELD & 7 & 13,708 & 9,989 & 1,109 & 2,610 & 12.2  & TV series & 3.2 / 305.0 & 8.7 / 72.0 & \multirow{2}{*}{English} \\
& IEMOCAP & 6 & 7,532 & 5,354 & 528 & 1,650 & 9.6 & Improvised scripts & 4.6 / 34.2 & 12.8 / 106.0 &  \\
\addlinespace[0.05cm]
\cline{1-12}
\addlinespace[0.05cm]
\multirow{2}{*}{Sentiment} 
& MOSI & 2 & 2,199 & 1,284 & 229 & 686  & 2.6 & Youtube & 4.3 / 52.5  & 12.5 / 114.0 & English\\
& CH-SIMS v2.0 & 3  & 4,403 & 2,722 & 647 & 1,034 & 4.3 & TV series, films & 3.6 / 42.7  & 1.8 / 7.0 & Mandarin\\ 
\addlinespace[0.05cm]
\cline{1-12}
\addlinespace[0.05cm]
\multirow{2}{*}{\makecell{Speaking \\ Style}}
& UR-FUNNY-v2 & 2 & 9,586 & 7,612 & 980 & 994 & 12.9 & TED & 4.8 / 325.7  & 16.3 / 126.0 & \multirow{2}{*}{English}  \\
& MUStARD & 2 & 690 & 414 & 138 & 138 & 1.0 & TV series & 5.2 / 20.0  & 13.1 / 68.0 &  \\ 
\addlinespace[0.05cm]
\cline{1-12}
\addlinespace[0.05cm]
\multirow{2}{*}{\makecell{Communication \\ Behavior}} 
& Anno-MI (client) & 3 & 4,713 & 3,123 & 461 & 1,129  & 10.8 & \multirow{2}{*}{\makecell{YouTube \\ \& Vimeo}} & 8.2 / 600.0  & 16.3 / 266.0 & \multirow{2}{*}{English}   \\ 
& Anno-MI (threapist) & 4 & 4,773 & 3,161 & 472 & 1,140  &  12.1 &  & 9.1 / 1316.1  & 17.9 / 205.0 &  \\ 
\bottomrule
\end{tabular}
}
\caption{\label{tab:datasets}
Dataset statistics for each dimension in the MMLA benchmark. \#C, \#U, \#Train, \#Val, and \#Test represent the number of label classes, utterances, training, validation, and testing samples, respectively. \textit{avg.} and \textit{max.} refer to the average and maximum lengths.}
\end{table*}

\subsection{Evaluation Dimensions}
\label{sec:evaluation_dimensions}
To comprehensively evaluate the complexity and diversity of human interactions, we select six representative dimensions across various linguistic and interactional levels: \textit{intent}, \textit{dialogue act}, \textit{emotion}, \textit{sentiment}, \textit{speaking style}, and \textit{communication behavior}. These dimensions collectively encapsulate the core aspects of multimodal language analysis~\cite{schroder2014intention,gandhi2023multimodal}. In particular, \textit{intent} captures the ultimate purpose or goal of human communication, such as requesting information or making decisions~\cite{ijcai2021-0622}. In contrast, \textit{dialogue act} is a more coarse-grained type of intent~\cite{saha2020towards}. It typically focuses on the dynamic progression of communication, such as questioning or stating opinions~\cite{stolcke2000dialogue}. Nonverbal signals (e.g., gaze shifts, gestures, and facial expressions) provide valuable clues to resolve ambiguities in both perspectives~\cite{mintrec,zhang2024mintrec2.0}.

\textit{Sentiment}, \textit{emotion}, and \textit{speaking style} are three significant aspects often accompanying communicative interactions. \textit{Sentiment} refers to the polarity (e.g., positive or negative) of subjective opinions~\cite{das2023multimodal}, \textit{emotion} conveys the speaker’s internal psychological state (e.g., happiness, anger)~\cite{ekman1992argument}, and \textit{speaking style} refers to individual expressive variations in communication (e.g., sarcasm, humor)~\cite{pennebaker2003psychological}. Multimodal cues (e.g., facial expressions and gestures) play a crucial role in inferring these communicative characteristics~\cite{yang2022disentangled,hasan2021humor,Patro_2021_WACV}. \textit{Communication behavior} explores the interaction behaviors between individuals (e.g., sustain, change, and reflection), which facilitate the progression of conversations and exhibit social properties within groups~\cite{Anno-MI}. Non-verbal signals (e.g., eye contact and gestures) can help uncover these behaviors and offer insights into modeling social cohesion~\cite{vinciarelli2009social}. Detailed information about the labels used for each dimension in each dataset can be found in Appendix~\ref{appendix:used_labels}.

\subsection{Data Sources}
We collect nine typical publicly available multimodal language datasets corresponding to the evaluation dimensions. Detailed statistics for these datasets are provided in Table~\ref{tab:datasets}. For the \textit{intent} dimension, we use two pioneering multimodal intent datasets, MIntRec~\cite{mintrec} and MIntRec2.0~\cite{zhang2024mintrec2.0}, which cover up to 30 intent classes commonly occurring in daily life. For the \textit{emotion} dimension, we utilize two widely used multimodal emotion recognition datasets, MELD~\cite{poria-etal-2019-meld} and IEMOCAP~\cite{busso2008iemocap}, both containing Ekman’s six universal emotion categories as suggested in~\cite{hu-etal-2022-unimse}. Additionally, MELD includes a \textit{neutral} class. For the \textit{dialogue act} dimension, we use curated versions of the MELD and IEMOCAP datasets, with annotations provided by EmoTyDA~\cite{saha2020towards}. These annotations consist of 12 commonly occurring classes selected from the SwitchBoard tag set~\cite{godfrey1992switchboard}. For the \textit{sentiment} dimension, we use two popular multimodal sentiment analysis datasets: MOSI~\cite{zadeh2016mosi} and CH-SIMS v2.0~\cite{liu2022make}. Both are annotated with sentiment intensity values in the range of [-3, 3]. Following~\cite{yu2021learning}, we convert these annotations into polarity-based two-class and three-class labels for evaluation. For the \textit{speaking style} dimension, we focus on two properties that play significant roles in social interactions: humor and sarcasm. We use UR-FUNNY-v2~\cite{hasan2019ur} and MUStARD~\cite{castro2019towards} for binary classification tasks, respectively. For the \textit{communication behavior} dimension, we employ the Anno-MI~\cite{Anno-MI} dataset, which involves motivational interviewing (MI) in counseling dialogues. This dataset is divided into two subsets, each analyzing three or four typical behaviors exhibited by clients and therapists. Details of the annotation quality assurance for each dataset are provided in Appendix~\ref{appendix:quality}.

These datasets contain a wide variety of characters, scenes, and background contexts in both English and Mandarin. They are sourced from popular TV series (e.g., \textit{Friends}, \textit{The Big Bang Theory}, \textit{Superstore}, etc.), films, online video-sharing platforms (e.g., YouTube, Vimeo, Bilibili), idea-sharing platforms (e.g., TED), and scripted dyadic sessions. We perform necessary cleaning and corrections to ensure the quality of each multimodal sample, aligning transcriptions, raw videos, and audio data. The datasets in the benchmark consist of 61,016 high-quality multimodal samples, totaling 76.6 hours of video, with 12,093 samples reserved for testing.

\subsection{Method Overview}
\label{sec:benchmark_method}

\textbf{Zero-shot Inference}. We leverage the generalization capabilities of foundation models for zero-shot inference. Specifically, for LLMs, the prompt template includes the transcribed utterances of speakers as text information, followed by a task-specific query with candidate labels. The LLM generates a response by predicting the next token in an autoregressive manner~\cite{brown2020language}, which corresponds to the most appropriate label. For MLLMs, we extend this template by adding the special token \texttt{<video>} at the beginning of the instruction, with its number aligned to the number of videos. This ensures structured alignment across modalities, enhancing the model’s capacity to process multimodal input. Details of the prompt templates used for inference can be found in Appendix~\ref{appendix:prompts}.

\textbf{Supervised Fine-tuning (SFT)}. We further optimize foundation models to enhance their instruction-following capabilities using SFT techniques while employing the same instruction templates as used during inference. Fine-tuning is performed by minimizing the cross-entropy loss between the model’s autoregressively predicted token probabilities and the ground-truth tokens corresponding to the labels. Let the input sequence be \( x = (x_1, x_2, \dots, x_n) \) and the target sequence be \( y = (y_1, y_2, \dots, y_m) \). The cross-entropy loss \( \mathcal{L}_{\text{CE}} \) is defined as:
\[
\mathcal{L}_{\text{CE}} = - \sum_{t=1}^{m} \log P(y_t | x, y_{<t}; \theta),
\]
where \( P(y_t | x, y_{<t}; \theta) \) is the probability of token \( y_t \) given the input \( x \), the previous tokens \( y_{<t} \), and the model parameters \( \theta \). To ensure training stability and reduce computational cost, we adopt the Low-Rank Adaptation (LoRA)~\cite{hu2022lora} technique, which  significantly reduces the number of parameters to be fine-tuned while preserving the model’s generalization capabilities.

\textbf{Instruction Tuning (IT)}. Since SFT addresses only the single-task scenario, we further explore the generalization ability of foundation models on multiple tasks. We first combine the training data from all datasets of each task for training, then we use the same template as the other two strategies, with the difference being that the task is not limited to one. The optimization objective follows SFT and uses the candidate labels of each task as supervised targets.

%% file: sec/4_experiments.tex
\section{Experiments}
\textbf{Evaluation Metrics}.
We employ six commonly used metrics: accuracy (ACC), weighted F1-score (WF1), weighted precision (WP), macro F1-score (F1), recall (R), and precision (P) for evaluation, as suggested in the literature~\cite{zhang2024mintrec2.0, poria-etal-2019-meld, zhou2024token, zadeh2016mosi, hasan2021humor}. In particular, we report the primary results of ACC in this paper, with additional results for the remaining metrics provided in the Appendices.

\textbf{Evaluation Baselines}. We apply the three evaluation methods as described in Section~\ref{sec:benchmark_method} on advanced LLMs and MLLMs as baselines. We also compare the foundation models with SOTA multimodal machine learning (MML) methods.
\begin{itemize}[left=0pt, itemsep=0pt] 
    \item \textbf{LLMs}. Three series of different parameter scales of unimodal foundation models are included: Llama-3~\cite{dubey2024llama} (8B), InternLM-2.5 (7B)~\cite{cai2024internlm2}, and Qwen2~\cite{yang2024qwen2} (0.5B, 1.5B, and 7B). 
    \item \textbf{MLLMs}. Five series of different parameter scales of multimodal foundation models are included: VideoLLaMA2~\cite{cheng2024videollama} (7B), Qwen2-VL~\cite{wang2024qwen2} (7B and 72B), LLaVA-Video~\cite{zhang2024video} (7B and 72B), LLaVA-OneVision (LLaVA-OV)~\cite{li2024llava} (7B and 72B), and MiniCPM-V-2.6~\cite{yao2024minicpm} (8B). For language decoding, the first series use Mistral~\cite{jiang2023mistral}, and the last four series use Qwen2 with the same parameter scale as the MLLM. We follow the same vision encoders as those in the corresponding released open-source models. We also apply zero-shot inference on one closed-source MLLM, GPT-4o~\cite{hurst2024gpt} as a baseline.
    \item \textbf{MML Methods}. We collect open-source MML methods with SOTA performance for each dataset for a detailed comparison. Specifically, for MIntRec: MIntOOD~\cite{zhang2024multimodal}, MIntRec2.0: MulT~\cite{tsai2019multimodal}, MELD and IEMOCAP: UniMSE~\cite{hu-etal-2022-unimse}, MELD-DA: TCL-MAP~\cite{zhou2024token}, IEMOCAP-DA: MIntOOD~\cite{zhang2024multimodal}, MOSI: MMML~\cite{wu2024multimodal}, CH-SIMS v2.0: ALMT~\cite{zhang2023learning}, UR-FUNNY-v2 and MUStARD: SimMMDG~\cite{dong2023simmmdg}. The results are reported as they appear in the corresponding papers.
\end{itemize}

\textbf{Experimental Setup}. We mostly follow the original data splits for training, validation, and testing for each dataset, as detailed in Table~\ref{tab:datasets}. Each sample consists of text and video data aligned at the utterance level for speakers. For SFT and IT methods, we utilize LLaMAFactory~\cite{zheng-etal-2024-llamafactory} for all LLMs and Qwen2-VL, SwiFT~\cite{zhao2025swift} for MiniCPM-V-2.6, LLaVA-NeXT~\footnote{https://github.com/LLaVA-VL/LLaVA-NeXT} for LLaVA-OV and LLaVA-Video, and VideoLLaMA2 using its own public code~\footnote{https://github.com/DAMO-NLP-SG/VideoLLaMA2}, respectively. We employ FlashAttention-2~\cite{dao2024flashattention} to optimize the attention modules of transformers, reducing memory and time costs. Besides, we leverage the DeepSpeed library for distributed training (e.g., using ZeRO-3 for memory optimization) and parallel computation. The precision type is set to BF16, offering reduced computational costs compared to FP16 or FP32. The learning rates range from 2e-5 to 1e-3, and a cosine learning rate scheduler with warmup ratios from 0.1 to 0.3 is applied. The training batch sizes are chosen from $\{4, 8, 16, 24\}$. The rank and $\alpha$ parameters of the LoRA module are set to $\{8, 16, 64, 128\}$ and $\{16, 32, 128, 256\}$, respectively. All experiments are conducted on NVIDIA A100 GPUs. We monitor model accuracy on the validation set to select the best checkpoint for inference. Details of the used hyperparameters and the full experimental results are shown in Appendix~\ref{appendix:results}.

%% file: sec/5_discussions.tex
\section{Results and Discussion}

\subsection{Main Results}

\begin{wrapfigure}{r}{0.36\textwidth} % "r" 表示右侧，宽度设为 0.4 文档宽度
\vspace{-0.4cm} % 调整此处的值
    \centering
    \includegraphics[width=1\linewidth]{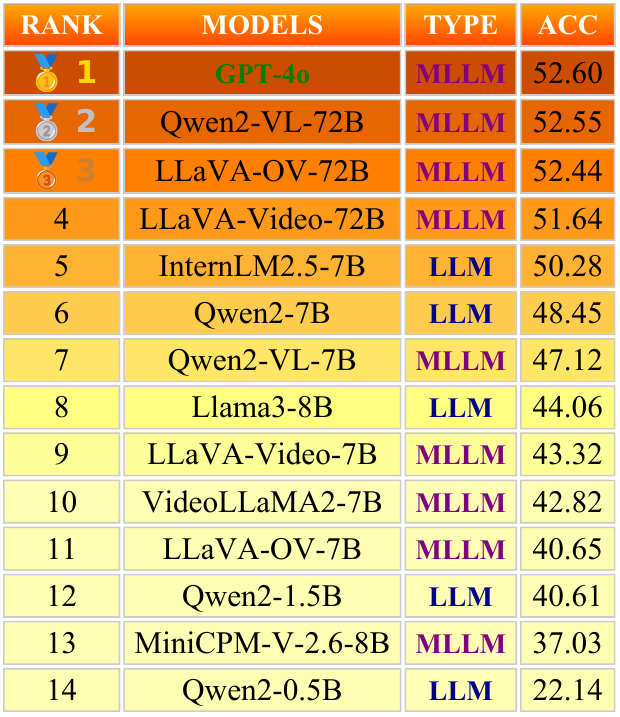}
    \caption{Rank of foundation models after zero-shot inference.}
    \label{fig:rank_infer}
\vspace{-1.0cm} % 调整此处的值
\end{wrapfigure}

\begin{wrapfigure}{r}{0.35\textwidth} % "r" 表示右侧，宽度设为 0.4 文档宽度
\vspace{-0.5cm} % 调整此处的值
    \centering
    \includegraphics[width=1\linewidth]{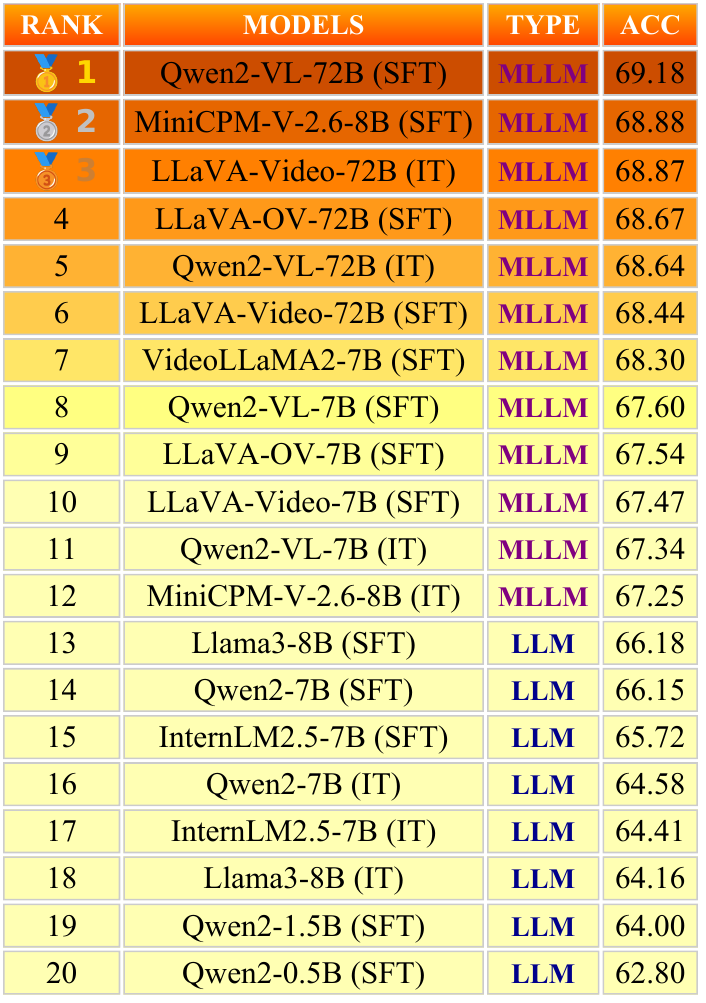}
    \caption{Rank of foundation models after SFT and IT.}
    \label{fig:rank_train}
    \vspace{-0.5cm} % 调整此处的值
\end{wrapfigure}

To clearly illustrate the performance differences between foundation models on the MMLA benchmark, we present the ranking statistics of the average accuracy (ACC) across all combined testing sets. Specifically, the zero-shot inference performance is shown in Figure~\ref{fig:rank_infer}, while the performance after SFT and IT is shown in Figure~\ref{fig:rank_train}. We find some interesting and new insights as below.

\textbf{Comparable Performance between LLMs and MLLMs in Zero-shot Inference}. As shown in Figure~\ref{fig:rank_infer}, the closed-source GPT-4o achieves the best performance, and the three open-source 72B MLLMs occupy the remaining positions in the top four. This is unsurprising, as these models contain more parameters and therefore exhibit stronger generalization and reasoning capabilities, consistent with the scaling laws~\cite{kaplan2020scaling}. However, we note that the much smaller-scale LLM InternLM2.5-7B achieves comparable performance, within approximately a 2\% difference. Furthermore, among models with a similar scale (7B or 8B parameters), most MLLMs exhibit lower performance than LLMs. For example, InternLM2.5 and Qwen2 outperform most MLLMs (e.g., LLaVA-Video, VideoLLaMA2) by 5$\sim$8\%. These results indicate that existing MLLMs have significant limitations in leveraging non-verbal information to capture complex high-level semantics without supervision from domain-specific data.

\textbf{Small MLLMs Rival Large Ones After SFT and IT.}  
Although MLLMs exhibit substantial performance gaps in zero‑shot inference, parameter size matters far less once they’re trained with SFT or IT. For example, as shown in Figure~\ref{fig:rank_train}, 7B MLLMs trained with SFT achieve 67.47$\sim$68.30\% ACC, while their 72B counterparts reach 68.44$\sim$69.18\%, a performance gap of only 1$\sim$2\%. Specially, the 8B MiniCPM-V-2.6 after SFT attains second place with 68.88\%, only 0.3\% lower in ACC than the top model, and surpasses several much larger MLLMs. 7B, 8B, and 72B MLLMs trained with IT also achieve ACC scores within 2\% of each other (i.e., 67.25$\sim$68.87\%). These results show that small‑scale well-trained MLLMs can capture the cognitive semantics underlying human language, suggesting lightweight foundation models are feasible and significantly reduce costs.

\textbf{Comparable Performance of MLLMs Between SFT and IT.} 
Although SFT offers the advantage of task-specific fine‑tuning to boost individual task performance, we observe that MLLMs trained via IT can achieve comparable results or even surpass those of SFT when evaluated on multiple tasks. For example, in Figure~\ref{fig:rank_train}, the 72B LLaVA‑Video ranks third and outperforms its SFT counterpart by 0.43\%. Qwen2‑VL shows only a slight performance drop after IT (0.54\% for 72B and 0.26\% for 7B). In contrast, MiniCPM‑V‑2.6 suffers a more pronounced decline of 1.63\% compared with its SFT counterpart, and some models (e.g., LLaVA‑Video and LLaVA‑OV) are omitted because they exhibit severe hallucinations, producing irrelevant outputs following IT. However, the strong performance of certain MLLMs highlights the potential of training a unified model to excel across diverse tasks without the overhead of maintaining multiple models, thereby demonstrating the robust generalization capabilities of MLLMs on this task.

\textbf{MLLMs Still Face Challenges on the MMLA Benchmark}. From Figures~\ref{fig:rank_infer} and~\ref{fig:rank_train}, we observe that the best MLLM in zero‑shot inference (GPT‑4o) achieves only 52.6\% ACC, and the best model after training with supervised data (72B Qwen2‑VL) reaches just 69.18\% ACC, still exhibiting huge limitations. These findings underscore the difficulty and importance of the MMLA benchmark, pushing the boundaries of existing MLLMs and laying a solid foundation for future related research.

\subsection{Fine-grained Performance on Different Dimensions}

\begin{figure*}[h]
    \centering
    \includegraphics[width=1\linewidth]{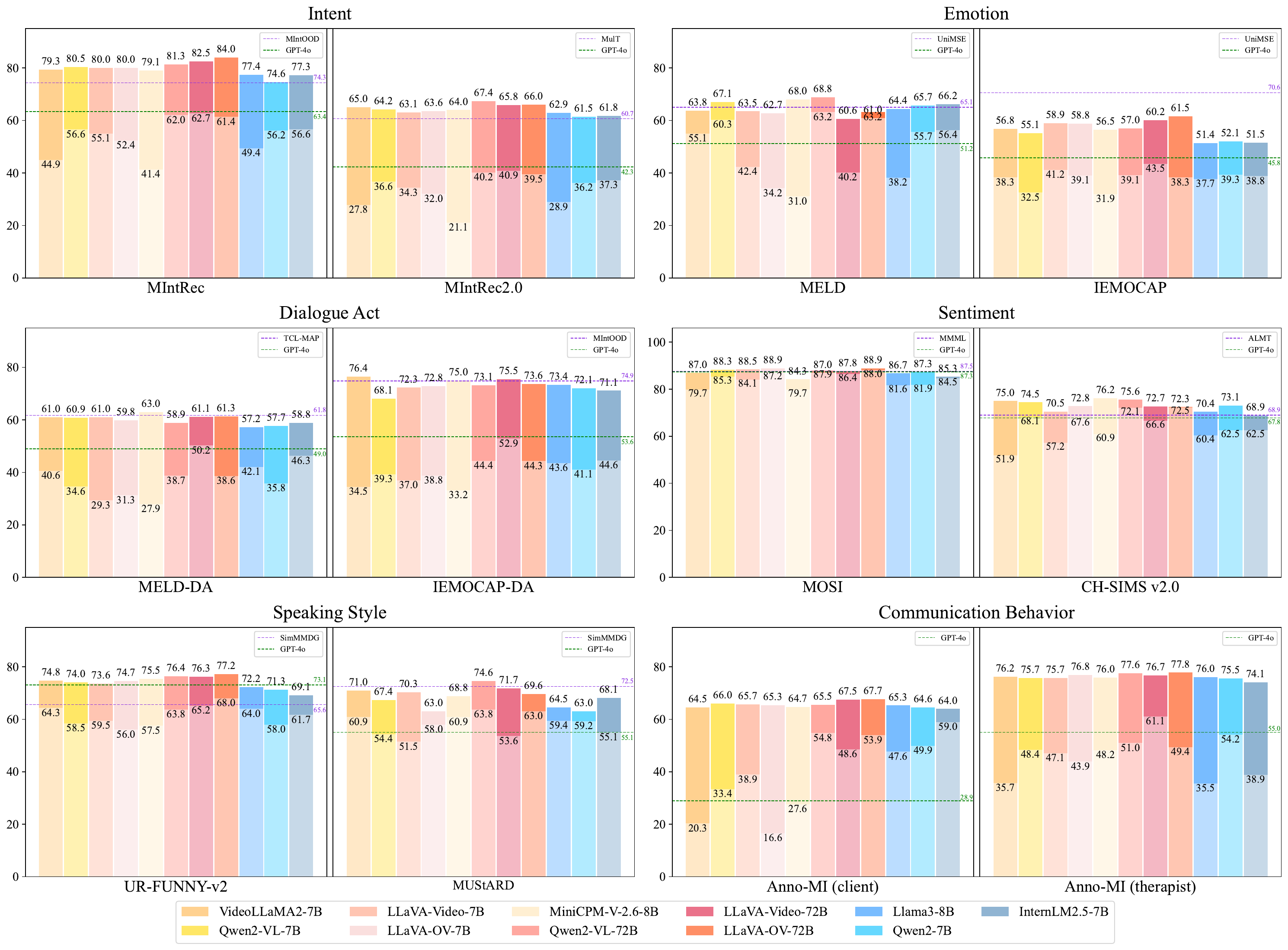}
    \caption{Fine‑grained zero‑shot inference and SFT performance (ACC). Within each bar, the light-colored lower segment corresponds to zero-shot inference performance, while the darker upper segment represents the additional gains from SFT. The performance of SOTA MML methods (if available) and GPT‑4o are indicated with purple and green dashed lines, respectively.}
    \label{fig:infer_sft_results}
    \vspace{-0.48cm} % 调整此处的值
\end{figure*}
\begin{figure*}[h]
    \centering
    \includegraphics[width=1\linewidth]{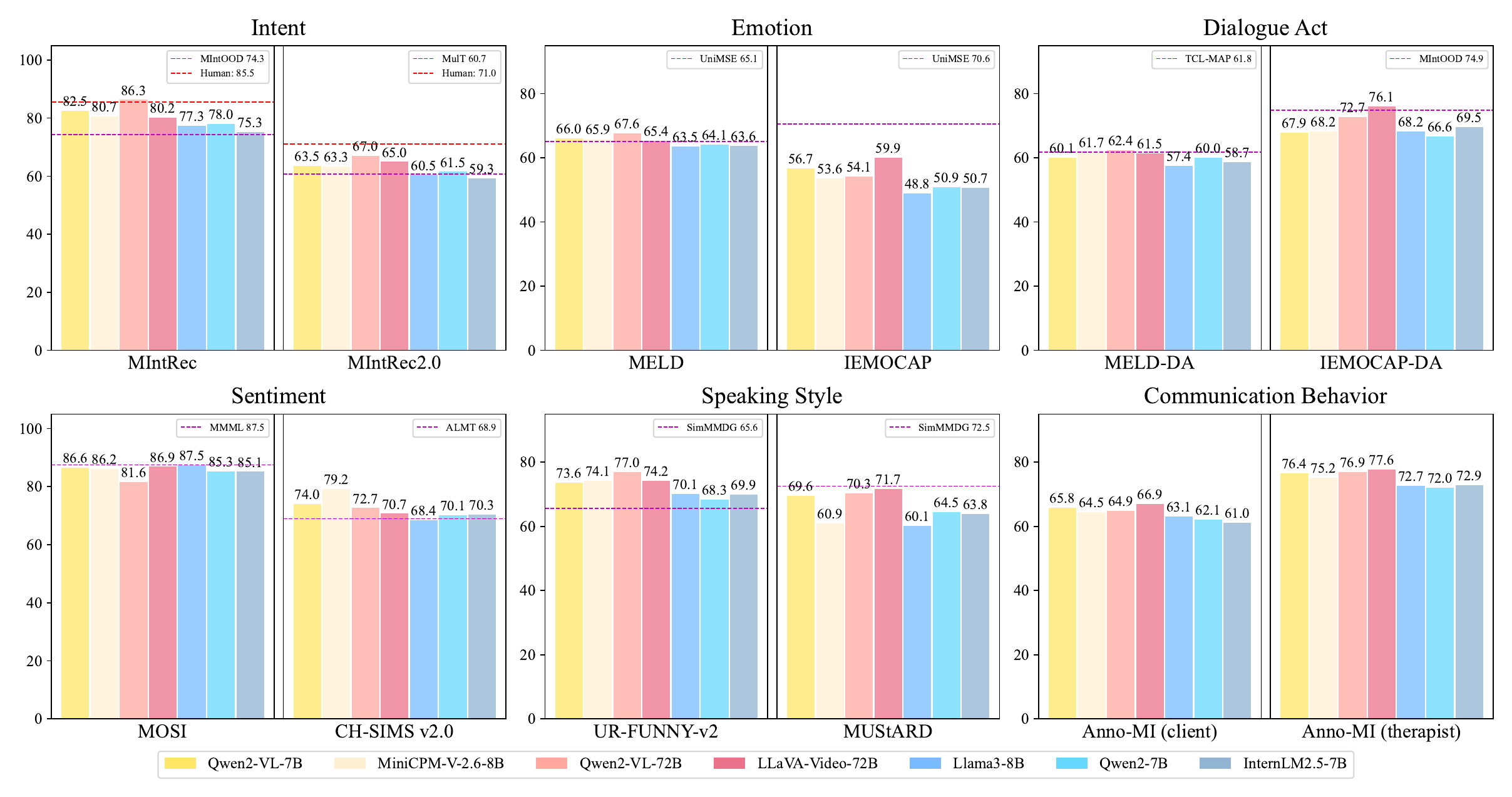}
    \caption{Fine‑grained performance (ACC) of instruction‑tuned MLLMs and LLMs on each dataset across six dimensions. The performance of SOTA MML methods and humans are indicated with dashed lines, if available.}
    \label{fig:instruct_results}
     \vspace{-0.7cm} % 调整此处的值
\end{figure*}
To further investigate fine‑grained performance across different dimensions, we present the results of three methods for each MLLM and LLM on every dataset, as shown in Figures~\ref{fig:infer_sft_results} and~\ref{fig:instruct_results}.

\textbf{Foundation Models Struggle with Zero-Shot Inference.} As shown in Figure~\ref{fig:infer_sft_results}, zero-shot performance is substantially limited, with ACC scores below 60\% on many challenging semantic dimensions (e.g., \textit{Intent}, \textit{Emotion}, \textit{Dialogue Act}, and \textit{Communication Behavior}). This shortcoming arises because these dimensions typically involve numerous categories with nuanced differences. In contrast, performance on the \textit{Sentiment} and \textit{Speaking Style} dimensions is generally higher because these tasks are simpler, requiring only two or three classes to be distinguished. GPT-4o achieves the best results in several dimensions, such as \textit{Intent}, \textit{Dialogue Act}, and \textit{Sentiment}, highlighting its strong ability to leverage multiple modalities for reasoning. However, it still struggles with tasks like sarcasm detection, emotion recognition, and communication behavior recognition, likely due to interference from scene context, background, and characters. Finally, while LLMs show performance comparable to or better than MLLMs of the same parameter scale, their scores remain below 60\% in most cases, underscoring the significant limitations of current foundation models on our benchmark.

\textbf{Foundation Models Significantly Improve After SFT.} As shown in Figure~\ref{fig:infer_sft_results}, foundation models exhibit a notable performance boost after SFT. For example, ACC scores increase by 20$\sim$40\% on \textit{Intent}, 10$\sim$40\% on \textit{Dialogue Act}, 4$\sim$20\% on \textit{Speaking Style}, and 5$\sim$50\% on \textit{Communication Behavior}. Specifically, MiniCPM-V-2.6 achieves improvements of over 30\% across most dimensions. These results demonstrate that training with supervised instruction data effectively helps MLLMs and LLMs distinguish complex semantic categories. Moreover, although both MLLMs and LLMs benefit from SFT, MLLMs consistently outperform LLMs (see Figure~\ref{fig:rank_train}), despite showing similar zero‑shot performance. This suggests that SFT not only aligns modalities better to activate multimodal reasoning, but also that incorporating non‑verbal information reduces hallucinations more effectively than using text alone. Finally, MLLMs after SFT set new state‑of‑the‑art results on most datasets except IEMOCAP and MUStARD, highlighting their great potential in multimodal language analysis.

% 第三段
\textbf{Foundation Models Master Multiple Tasks After IT.} As shown in Figure~\ref{fig:instruct_results}, MLLMs after IT can simultaneously match or surpass previous SOTA methods on most datasets. In particular, \textit{72B Qwen2‑VL is the first to exceed human performance on MIntRec~\cite{mintrec} (86.3\% vs.\ 85.5\%)}, marking remarkable progress toward human‑level semantic comprehension. 72B LLaVA‑Video improves over the SOTA method by 6.3\% and approaches human performance on MIntRec2.0~\cite{zhang2024mintrec2.0}. Similarly, most MLLMs exhibit superior results on sentiment analysis (Ch‑sims‑v2), humor detection (UR‑FUNNY‑v2), and emotion recognition (MELD). We also observe that the small‑scale MLLM (i.e., 8B MiniCPM‑V‑2.6) outperforms SOTA on seven datasets across five dimensions and achieves the best score on Ch‑sims‑v2. Moreover, small‑scale MLLMs outperform LLMs on nearly every dataset and task, underscoring that IT enhances multimodal reasoning and demonstrating the potential of training a unified MLLM to tackle multiple complex multimodal language tasks.

\subsection{Scalability of Foundation Models on MMLA}

\begin{wrapfigure}{r}{0.42\textwidth} % "r" 表示右侧，宽度设为 0.4 文档宽度
\vspace{-0.4cm} % 调整此处的值
    \centering
    \includegraphics[width=1\linewidth]{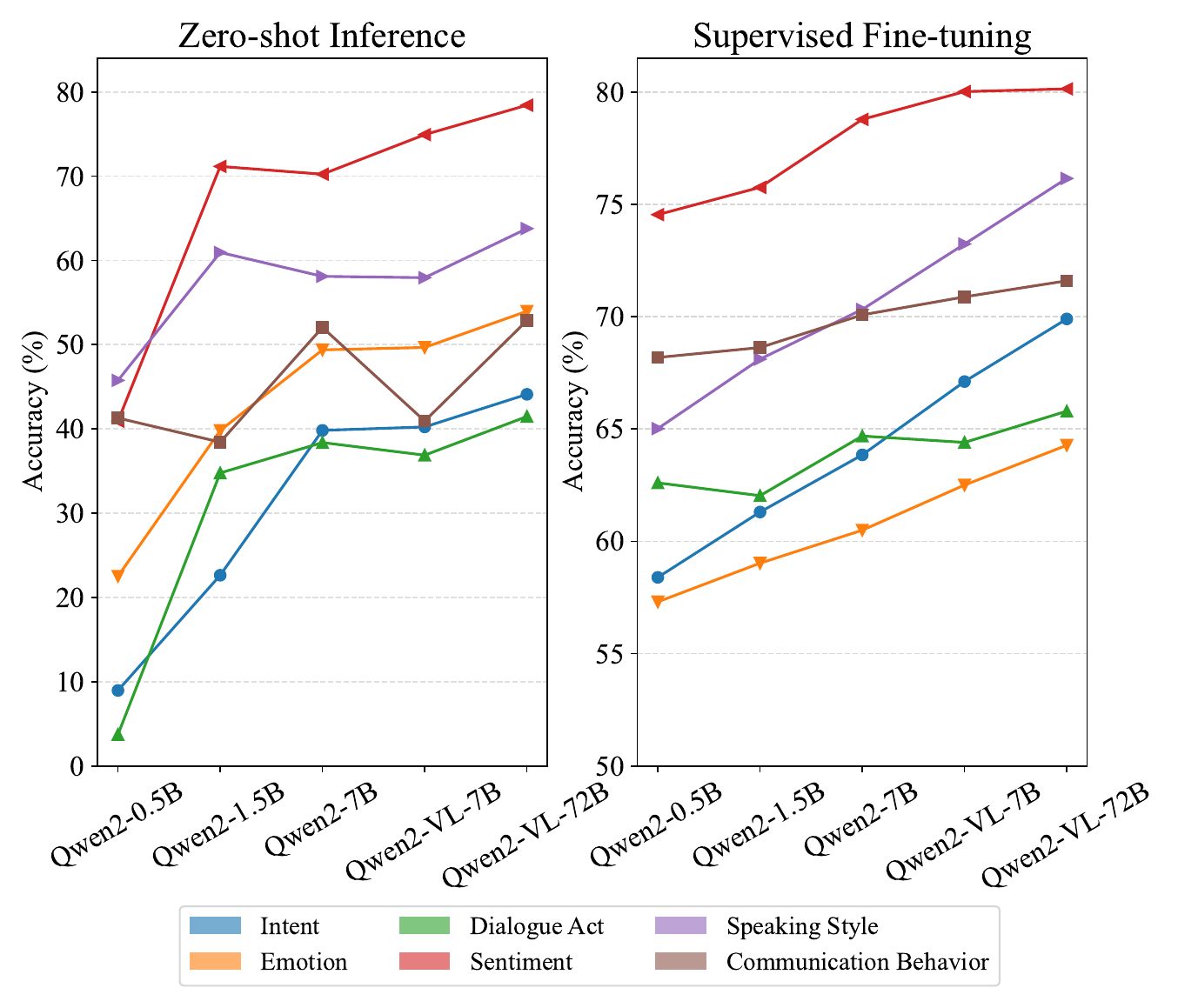}
    \caption{Scalability of Qwen2 and Qwen2-VL on the MMLA benchmark.}
    \label{fig:Scalability}
\vspace{-0.4cm} % 调整此处的值
\end{wrapfigure}

To examine the scalability of foundation models~\cite{kaplan2020scaling}, we analyze the effect of parameter scale using Qwen2 and Qwen2‑VL, presenting both zero‑shot inference and SFT results in Figure~\ref{fig:Scalability}.

\textbf{Scaling Performance of Zero‑Shot Inference.} In zero‑shot inference, scaling Qwen2 from 0.5B to 1.5B parameters achieves significant improvements across all dimensions except \textit{Communication Behavior}. When scaling from 1.5B to 7B, performance gains accelerate on \textit{Intent} and \textit{Communication Behavior}, slow down on \textit{Emotion} and \textit{Dialogue Act}, and even slightly decrease on \textit{Sentiment} and \textit{Speaking Style}. This phenomenon indicates that larger gains occur with smaller scale changes. When moving from Qwen2 to Qwen2‑VL, performance is comparable or better in all dimensions except for \textit{Communication Behavior}, which shows a dramatic drop. However, scaling Qwen2‑VL from 7B to 72B yields substantial improvements, further validating the scalability of MLLMs.

\textbf{Scaling Performance of SFT}. After SFT, scaling Qwen2 from 0.5B to 7B yields modest improvements of 3$\sim$5\% on the \textit{Intent}, \textit{Sentiment}, \textit{Speaking Style}, and \textit{Emotion} dimensions, with limited gains of less than 2\% on \textit{Communication Behavior} and \textit{Dialogue Act}. Besides, scaling Qwen2‑VL from 7B to 72B achieves substantial improvements of over 5\% on \textit{Speaking Style} and \textit{Intent} dimensions, while yielding under 2\% gains in \textit{Sentiment}, \textit{Communication Behavior}, and \textit{Dialogue Act}. These results suggest that simply enlarging model parameters provides little benefit for analyzing complex multimodal language semantics when using supervised instructions as prior knowledge. They also highlight the significant challenge posed by this benchmark and underscore the need to design appropriate architectures and curate high-quality data for learning high-level cognitive semantics.

%% file: sec/6_conclusions.tex
\section{Conclusions}
This paper proposes MMLA, the first large-scale benchmark for evaluating foundation models on multimodal language analysis. It covers six core semantic dimensions across more than 61,000 utterances from nine diverse datasets spanning text, audio, and video modalities. We evaluate five branches of MLLMs and three branches of LLMs, ranging from 0.5B to 72B parameters, using three methods to provide a comprehensive analysis. This benchmark yields several new insights. First, existing MLLMs exhibit poor capabilities and offer no advantage over LLMs in zero-shot inference. Second, supervised fine-tuning (SFT) effectively activates MLLMs, enabling them to leverage non-verbal modalities to understand cognitive-level semantics, and achieves substantial improvements over LLMs. Third, instruction tuning (IT) can further fine-tune a unified model to achieve comparable or better performance on all SFT tasks. Interestingly, we find that smaller MLLMs, after both SFT and IT, demonstrate enormous potential, achieving performance comparable to much larger models while significantly reducing computational costs. Finally, existing MLLMs still face significant challenges, with an average accuracy below 70\%, underscoring the importance and difficulty of the proposed benchmark and pushing the limits of research in multimodal language analysis. MMLA establishes a rigorous foundation for advancing multimodal language understanding and cognitive-level human–AI interaction.

%% file: sec/supplementary.tex
\clearpage
\appendix

\section{License}
\label{appendix:license}
This benchmark uses nine datasets, each of which is employed strictly in accordance with its official license and exclusively for academic research purposes. We fully respect the datasets’ copyright policies, license requirements, and ethical standards. For those datasets whose licenses explicitly permit redistribution, we release the original video data (e.g., MIntRec\footnote{\url{https://github.com/thuiar/MIntRec}}, MIntRec2.0\footnote{\url{https://github.com/thuiar/MIntRec2.0}}, MELD\footnote{\url{https://github.com/declare-lab/MELD}}, UR-FUNNY-v2\footnote{\url{https://github.com/ROC-HCI/UR-FUNNY}}, MUStARD\footnote{\url{https://github.com/soujanyaporia/MUStARD}}, MELD-DA and IEMOCAP-DA\footnote{\url{https://github.com/sahatulika15/EMOTyDA}}, CH-SIMS v2.0\footnote{\url{https://github.com/thuiar/ch-sims-v2}}). For datasets that restrict video redistribution, users should obtain the videos directly from their official repositories (e.g., MOSI\footnote{\url{https://github.com/matsuolab/CMU-MultimodalSDK}}, IEMOCAP\footnote{\url{https://sail.usc.edu/iemocap}}, Anno-MI\footnote{\url{https://github.com/uccollab/AnnoMI}}). In compliance with all relevant licenses, we also provide the original textual data unchanged, together with the specific dataset splits used in our experiments. This approach ensures reproducibility and academic transparency while strictly adhering to copyright obligations and protecting the privacy of individuals featured in the videos.

\section{Used Labels for Each Dataset}
\label{appendix:used_labels}
\begin{itemize}[left=0pt, itemsep=0pt]
    \item \textbf{Intent}. The MIntRec dataset uses 20 predefined intent categories derived from two coarse‑grained classes (i.e., \textit{achieve goals} and \textit{express emotions and attitudes}), as described in~\cite{mintrec}. These categories are: \textit{complain}, \textit{praise}, \textit{apologize}, \textit{thank}, \textit{criticize}, \textit{agree}, \textit{taunt}, \textit{flaunt}, \textit{joke}, \textit{oppose}, \textit{comfort}, \textit{care}, \textit{inform}, \textit{advise}, \textit{arrange}, \textit{introduce}, \textit{leave}, \textit{prevent}, \textit{greet}, and \textit{ask for help}. MIntRec2.0 adds 10 more labels (i.e., \textit{doubt}, \textit{acknowledge}, \textit{refuse}, \textit{warn}, \textit{emphasize}, \textit{ask for opinions}, \textit{confirm}, \textit{explain}, \textit{invite}, and \textit{plan}) to the original 20. We use the in‑scope portion of this dataset for intent recognition.

    \item \textbf{Dialogue Act}. The MELD‑DA and IEMOCAP‑DA datasets select the 12 most frequent dialogue‑act tags in everyday conversation, based on the 42 acts defined in~\cite{stolcke2000dialogue}. The chosen tags are: \textit{greeting}, \textit{question}, \textit{answer}, \textit{statement‑opinion}, \textit{statement‑non‑opinion}, \textit{apology}, \textit{command}, \textit{agreement}, \textit{disagreement}, \textit{acknowledge}, \textit{backchannel}, and \textit{others}.

    \item \textbf{Emotion}. The IEMOCAP dataset adopts Ekman’s six universal emotions (as in prior work~\cite{wang2019words, hu2022mm}): \textit{angry}, \textit{happy}, \textit{sad}, \textit{neutral}, \textit{frustrated}, and \textit{excited}. The MELD dataset uses seven emotion classes~\cite{poria-etal-2019-meld}: \textit{neutral}, \textit{surprise}, \textit{fear}, \textit{sadness}, \textit{joy}, \textit{anger}, and \textit{disgust}.

    \item \textbf{Sentiment}. For the MOSI and CH‑SIMS v2.0 datasets, sentiment intensity scores range from \(-3\) to \(3\) and are mapped to two‑ or three‑way polarity classes (e.g., \textit{positive}, \textit{neutral}, \textit{negative}), as recommended in~\cite{zadeh2016mosi, liu2022make}.

    \item \textbf{Speaking Style}. The UR‑FUNNY‑v2 and MUStARD datasets both perform binary classification tasks: humor detection (\textit{humorous} vs.\ \textit{serious}) and sarcasm detection (\textit{sarcastic} vs.\ \textit{sincere}), respectively.

    \item \textbf{Communication Behavior}. The Anno‑MI dataset is split into two parts for counseling dialogue analysis. The first part contains four therapist communication skills: \textit{question}, \textit{input}, \textit{reflection}, and \textit{other}. The second part contains three client talk types: \textit{change}, \textit{neutral}, and \textit{sustain}.
\end{itemize}

\section{Assurance of Annotation Quality}
\label{appendix:quality}
We employ rigorous procedures to select datasets with high-quality annotations. Quality is ensured through the following strategies and statistical measures for each dataset:

\begin{itemize}[left=0pt, itemsep=0pt]
    \item \textbf{MIntRec}~\cite{mintrec} and \textbf{MIntRec2.0}~\cite{zhang2024mintrec2.0}. Intent labels are assigned by majority voting (three of five and two of three annotators, respectively). Fleiss’s kappa values of 0.88 for MIntRec and 0.69 for MIntRec2.0 indicate excellent and substantial agreement, respectively.
    \item \textbf{MELD}~\cite{poria-etal-2019-meld} and \textbf{IEMOCAP}~\cite{busso2008iemocap}. Emotion labels are determined by three-annotator majority voting, yielding Fleiss’s kappa values of 0.43 (MELD) and 0.40 (IEMOCAP), reflecting acceptable reliability for emotion annotation.
    \item \textbf{MELD-DA} and \textbf{IEMOCAP-DA}~\cite{saha2020towards}. Dialogue-act labels are annotated by three experts, achieving over 80\% inter-annotator agreement.
    \item \textbf{MUStARD}~\cite{castro2019towards}. Three annotators achieved a Cohen’s kappa of 0.588 for sarcasm detection.
    \item \textbf{UR-FUNNY-v2}~\cite{hasan2019ur}. The original UR-FUNNY was annotated based on direct laughter markers in punchlines; noisy and overlapping instances were removed to form the second version, which we use in our benchmark.
    \item \textbf{MOSI}~\cite{zadeh2016mosi}. Five master workers (approval rate > 95\%) annotated sentiment intensity, with a Krippendorff’s alpha of 0.77.
    \item \textbf{Anno-MI}~\cite{Anno-MI}. Ten therapists from the International Organization of Authoritative Motivational-Interviewing Trainers annotated communication behavior labels, with Fleiss’s kappa values of 0.74 (therapist) and 0.47 (client), indicating substantial and moderate agreement, respectively.
    \item \textbf{CH-SIMS v2.0}~\cite{liu2022make}. This version corrects potential errors in the original CH-SIMS~\cite{yu-etal-2020-ch}. Seven well-trained annotators rated sentiment intensity: the highest and lowest scores were removed, and the average of the remaining five was mapped to discrete sentiment labels. We use this latest release, which also addresses potential misalignment issues.
\end{itemize}

\section{Used Prompts}
Zero-shot inference, supervised fine-tuning (SFT), and instruction tuning (IT) employ the following template:
\label{appendix:prompts}

\begin{alltt}
You are presented with a video in which the speaker says: {\textit{<context>}}. 
Based on the textual, visual, and audio content, what is the {\textit{<dimension>}}
of this speaker? 
The candidate labels for {\textit{<dimension>}} are: {\textit{<the list of labels>}}. 
Respond in the following format: {\textit{<dimension>}}: {\textit{label}}.
Only one label should be provided.
\end{alltt}
Here, \textit{<context>} denotes the the speaker's utterance, while \textit{<dimension>} refers to one of the six evaluation dimensions described in Section~\ref{sec:evaluation_dimensions}. The placeholder \textit{<the list of labels>} corresponds to the specific label set for each dimension (cf.\ Appendix~\ref{appendix:used_labels}). In SFT, the ground‑truth dimension and label are provided for supervised training. In IT, neither the queried dimension nor its label set is fixed to a single dataset, unlike in SFT.

\section{Detailed Experimental Results}
\label{appendix:results}
Due to space constraints, the main paper presents only a subset of the results. Here, we provide the complete results for zero-shot inference, supervised fine-tuning, and instruction-tuning across six evaluation metrics (ACC, WF1, WP, F1, P, R) in Table~\ref{tab:results-zero}, Table~\ref{tab:results-sft}, and Table~\ref{tab:results-it}. For both LLMs and MLLMs, the best and second-best results are highlighted in bold and underline, respectively.

The results align with the discussions and conclusions in the main paper. For zero-shot inference, multimodal models with 72B parameters achieve the best overall performance across all datasets. However, when comparing LLMs and MLLMs of the same scale, LLMs often exhibit competitive or even superior performance. After supervised fine-tuning, MLLMs show significant improvements and surpass LLMs on almost all datasets across all six metrics, underscoring the importance of incorporating non-verbal modalities for cognitively demanding tasks. After instruction-tuning, both the 7B and 72B MLLMs achieve excellent performance on all tasks, with results comparable to or better than those from supervised fine-tuning, indicating the potential of small-scale MLLMs to solve multiple tasks simultaneously. Moreover, under this evaluation protocol, MLLMs also outperform LLMs, further confirming the benefit of leveraging non-verbal modalities.

We also evaluate one powerful closed-source MLLM, GPT-4o. Specifically, we use OpenAI’s GPT-4o API for zero-shot inference with the same prompts as in Appendix~\ref{appendix:prompts}. During inference, we find that GPT-4o can be overly cautious with certain videos. For example, it sometimes fails to select a label from the candidate list and instead outputs responses like: \textit{I'm unable to determine the {\textit{dimension}} based on the given information}. To address this, we iteratively modify the prompts based on such outputs until GPT-4o consistently chooses a label from the list. The final results, shown in Table~\ref{tab:results-zero}, demonstrate that GPT-4o achieves the best or second-best performance on most metrics across datasets, highlighting its effectiveness on this challenging task.

Details of the hyperparameters used for supervised fine-tuning (SFT) and instruction-tuning (IT) of all LLMs and MLLMs are provided in Table~\ref{tab:param-1}, Table~\ref{tab:param-2}, Table~\ref{tab:param-3}, and Table~\ref{tab:param-4}.

\begin{table*}[h] 
    \scriptsize
    \tiny
    \centering
    \setlength{\tabcolsep}{2.5pt}
    \caption{\label{tab:results-zero} Full experimental results on the MMLA benchmark using zero-shot inference.}
    \scalebox{0.86}{
    \vspace{2mm}
    \centering
    \setlength{\belowcaptionskip}{-3.7cm}
\begin{tabular}{l|c|>
    {\centering\arraybackslash}p{2.1cm}|>
    {\centering\arraybackslash}p{2.1cm}|>{\centering\arraybackslash}p{2.1cm}|>{\centering\arraybackslash}p{2.1cm}|>{\centering\arraybackslash}p{2.1cm}|>{\centering\arraybackslash}p{2.1cm}|>{\centering\arraybackslash}p{2.1cm}|}
        \toprule
        \addlinespace[0.1cm]
         & Datasets & \multicolumn{2}{c|}{MIntRec} & \multicolumn{2}{c|}{MIntRec2.0} & \multicolumn{2}{c|}{MELD}\\
         \addlinespace[0.1cm]
          \cline{2-8}
          \addlinespace[0.1cm]
        
         & Models 
         & ACC \ \hspace{0.03cm} WF1 \ \hspace{0.03cm}  WP  \hspace{0.03cm} 
         & F1 \ \hspace{0.21cm} R \ \hspace{0.21cm}  P \hspace{0.03cm}
         & ACC \ \hspace{0.03cm} WF1 \ \hspace{0.03cm}  WP \hspace{0.03cm} 
         & F1 \ \hspace{0.21cm} R \ \hspace{0.21cm}  P \hspace{0.03cm}
         & ACC \ \hspace{0.03cm} WF1 \ \hspace{0.03cm}  WP  \hspace{0.03cm} 
         & F1 \ \hspace{0.21cm} R \ \hspace{0.21cm}  P \hspace{0.03cm} \\

         \midrule
         \multirow{8}{*}{\rotatebox[origin=c]{90}{LLMs}}
            & Qwen2-0.5B & 18.88 \ 15.08 \ 26.81 & 9.03 \ 10.03 \ 16.06 & 6.79 \ 4.08 \ 11.33 & 3.15 \ 5.49 \ 9.48 & 21.03 \ 12.89 \ 52.20 & 11.79 \ 19.00 \ 37.51 \\
            & Llama-3.2-1B & 23.60 \ 18.60 \ 26.71 & 15.08 \ 16.06 \ 25.55 & 13.63 \ 9.93 \ 23.98 & 9.56 \ 12.98 \ 22.73 & 22.34 \ 18.60 \ 47.81 & 18.30 \ 23.33 \ 25.34 \\
            & Qwen2-1.5B & 33.26 \ 24.79 \ 29.97 & 17.29 \ 21.16 \ 21.48 & 20.31 \ 16.59 \ 34.48 & 15.36 \ 18.79 \ 28.24 & 43.56 \ 45.35 \ 51.16 & 30.97 \ 33.85 \ 32.05 \\
            & Llama-3.2-3B & 37.98 \ 33.62 \ 42.04 & 30.99 \ 35.60 \ 36.84 & 28.58 \ 26.34 \ 42.81 & 23.27 \ 26.76 \ 34.87 & 38.24 \ 38.56 \ 59.46 & 30.60 \ 34.59 \ 34.35 \\
\bottomrule
    \end{tabular}
    }
    \scalebox{0.86}{
    \begin{tabular}{l|c|>
    {\centering\arraybackslash}p{2.1cm}|>
    {\centering\arraybackslash}p{2.1cm}|>{\centering\arraybackslash}p{2.1cm}|>{\centering\arraybackslash}p{2.1cm}|>{\centering\arraybackslash}p{2.1cm}|>{\centering\arraybackslash}p{2.1cm}|>{\centering\arraybackslash}p{2.1cm}|}
        \toprule
        \addlinespace[0.1cm]
         & Datasets & \multicolumn{2}{c|}{IEMOCAP} & \multicolumn{2}{c|}{MELD-DA} & \multicolumn{2}{c|}{IEMOCAP-DA}\\
                   \cline{2-8}
          \addlinespace[0.1cm]
         & Models 
         & ACC \ \hspace{0.03cm} WF1 \ \hspace{0.03cm}  WP  \hspace{0.03cm} 
         & F1 \ \hspace{0.21cm} R \ \hspace{0.21cm}  P \hspace{0.03cm}
         & ACC \ \hspace{0.03cm} WF1 \ \hspace{0.03cm}  WP \hspace{0.03cm} 
         & F1 \ \hspace{0.21cm} R \ \hspace{0.21cm}  P \hspace{0.03cm}
         & ACC \ \hspace{0.03cm} WF1 \ \hspace{0.03cm}  WP  \hspace{0.03cm} 
         & F1 \ \hspace{0.21cm} R \ \hspace{0.21cm}  P \hspace{0.03cm} \\

         \midrule

         \multirow{8}{*}{\rotatebox[origin=c]{90}{LLMs}}
            & Qwen2-0.5B & 24.91 \ 12.40 \ 41.94 & 9.53 \ 16.82 \ 30.62 & 4.75 \ 2.06 \ 8.00 & 4.31 \ 8.87 \ 7.28 & 2.65 \ 2.92 \ 19.23 & 4.49 \ 13.73 \ 9.53 \\
            & Llama-3.2-1B & 24.17 \ 23.21 \ 41.64 & 19.17 \ 21.45 \ 35.61 & 11.41 \ 8.74 \ 13.69 & 9.86 \ 13.01 \ 12.93 & 18.21 \ 13.09 \ 16.81 & 10.74 \ 18.45 \ 12.60 \\
            & Qwen2-1.5B & 33.79 \ 29.16 \ \underline{49.46} & 24.17 \ 26.56 \ 41.62 & 31.43 \ 27.45 \ 37.87 & 21.87 \ 27.10 \ 25.37 & 38.32 \ 32.71 \ 35.87 & 21.93 \ 26.21 \ 22.16 \\
            & Llama-3.2-3B & 35.02 \ 35.47 \ 42.31 & 29.34 \ 28.98 \ 35.67 & 33.08 \ 33.65 \ 45.47 & 25.61 \ 34.62 \ 28.06 & 33.92 \ 34.88 \ 44.73 & 21.81 \ 31.54 \ 23.71 \\
            & Qwen2-7B & \textbf{39.27} \ \underline{37.35} \ 48.95 & 31.11 \ \underline{31.54} \ \underline{41.75} & 35.84 \ 35.77 \ 46.81 & \underline{29.72} \ 35.78 \ \underline{33.05} & 41.08 \ 40.13 \ 50.25 & \underline{28.92} \ 33.65 \ 32.85 \\
\bottomrule
    \end{tabular}
    }
     \scalebox{0.86}{
    \begin{tabular}{l|c|>
    {\centering\arraybackslash}p{2.1cm}|>
    {\centering\arraybackslash}p{2.1cm}|>{\centering\arraybackslash}p{2.1cm}|>{\centering\arraybackslash}p{2.1cm}|>{\centering\arraybackslash}p{2.1cm}|>{\centering\arraybackslash}p{2.1cm}|>{\centering\arraybackslash}p{2.1cm}|}
        \toprule
        \addlinespace[0.1cm]
        & Datasets & \multicolumn{2}{c|}{MOSI} & \multicolumn{2}{c|}{CH-SIMS v2.0} & \multicolumn{2}{c|}{UR-FUNNY-v2}\\
                   \cline{2-8}
          \addlinespace[0.1cm]
         & Models 
         & ACC \ \hspace{0.03cm} WF1 \ \hspace{0.03cm}  WP  \hspace{0.03cm} 
         & F1 \ \hspace{0.21cm} R \ \hspace{0.21cm}  P \hspace{0.03cm}
         & ACC \ \hspace{0.03cm} WF1 \ \hspace{0.03cm}  WP \hspace{0.03cm} 
         & F1 \ \hspace{0.21cm} R \ \hspace{0.21cm}  P \hspace{0.03cm}
         & ACC \ \hspace{0.03cm} WF1 \ \hspace{0.03cm}  WP  \hspace{0.03cm} 
         & F1 \ \hspace{0.21cm} R \ \hspace{0.21cm}  P \hspace{0.03cm} \\

         \midrule

         \multirow{8}{*}{\rotatebox[origin=c]{90}{LLMs}}
            & Qwen2-0.5B & 61.22 \ 67.27 \ 74.76 & 44.83 \ 40.79 \ 49.84 & 27.47 \ 25.21 \ 68.65 & 25.24 \ 38.77 \ \underline{54.10} & 45.27 \ 47.11 \ 49.27 & 31.42 \ 30.21 \ 32.84 \\
            & Llama-3.2-1B & 70.99 \ 72.57 \ 78.10 & 48.36 \ 47.23 \ 52.15 & 65.28 \ 61.12 \ 58.60 & 34.55 \ 36.63 \ 33.36 & 55.43 \ 55.08 \ 55.48 & 55.02 \ 55.31 \ 55.48 \\
            & Qwen2-1.5B & 80.17 \ 81.50 \ 84.15 & 54.35 \ 53.52 \ 56.06 & 65.18 \ \underline{66.94} \ 69.64 & 43.02 \ 43.75 \ 43.45 & 62.37 \ 61.23 \ 63.93 & 40.75 \ 41.42 \ 42.66 \\
            & Llama-3.2-3B & 81.20 \ 82.52 \ 86.63 & 54.98 \ 54.03 \ 57.81 & \underline{66.73} \ 65.28 \ 64.29 & 39.44 \ 39.79 \ 39.78 & \underline{66.80} \ \underline{66.79} \ 66.80 & \textbf{66.78} \ \textbf{66.78} \ \textbf{66.80} \\
\bottomrule
    \end{tabular}
    }
     \scalebox{0.86}{
    \begin{tabular}{l|c|>
    {\centering\arraybackslash}p{2.1cm}|>
    {\centering\arraybackslash}p{2.1cm}|>{\centering\arraybackslash}p{2.1cm}|>{\centering\arraybackslash}p{2.1cm}|>{\centering\arraybackslash}p{2.1cm}|>{\centering\arraybackslash}p{2.1cm}|>{\centering\arraybackslash}p{2.1cm}|}
        \toprule
        \addlinespace[0.1cm]
        & Datasets & \multicolumn{2}{c|}{MUStARD} & \multicolumn{2}{c|}{Anno-MI (client)} & \multicolumn{2}{c|}{Anno-MI (therapist)}\\
                   \cline{2-8}
          \addlinespace[0.1cm]
         & Models 
         & ACC \ \hspace{0.03cm} WF1 \ \hspace{0.03cm}  WP  \hspace{0.03cm} 
         & F1 \ \hspace{0.21cm} R \ \hspace{0.21cm}  P \hspace{0.03cm}
         & ACC \ \hspace{0.03cm} WF1 \ \hspace{0.03cm}  WP \hspace{0.03cm} 
         & F1 \ \hspace{0.21cm} R \ \hspace{0.21cm}  P \hspace{0.03cm}
         & ACC \ \hspace{0.03cm} WF1 \ \hspace{0.03cm}  WP  \hspace{0.03cm} 
         & F1 \ \hspace{0.21cm} R \ \hspace{0.21cm}  P \hspace{0.03cm} \\

         \midrule

         \multirow{8}{*}{\rotatebox[origin=c]{90}{LLMs}}
            & Qwen2-0.5B & 50.00 \ 33.50 \ 25.18 & 22.33 \ 33.33 \ 16.79 & \underline{57.04} \ 49.22 \ 49.63 & 25.14 \ 27.58 \ 31.72 & 25.70 \ 25.55 \ 27.59 & 18.76 \ 19.27 \ 19.71 \\
            & Llama-3.2-1B & 51.45 \ 37.61 \ \underline{62.87} & 37.61 \ 51.45 \ \textbf{62.87} & 28.34 \ 30.10 \ 45.86 & 26.30 \ \underline{34.79} \ \underline{34.64} & 22.19 \ 22.89 \ 30.10 & 16.97 \ 16.70 \ 21.86 \\
            & Qwen2-1.5B & 52.90 \ 52.90 \ 52.90 & \underline{52.90} \ 52.90 \ 52.90 & 36.67 \ 40.45 \ 52.43 & 25.72 \ 28.48 \ 30.84 & 40.18 \ 37.14 \ 47.12 & 37.73 \ 41.17 \ 45.99 \\
            & Llama-3.2-3B & \underline{57.25} \ 52.16 \ 62.61 & 52.16 \ \underline{57.25} \ \underline{62.61} & 43.93 \ 46.77 \ 52.03 & 28.33 \ 26.70 \ 32.97 & 43.60 \ 43.35 \ 49.07 & 34.43 \ 34.98 \ 38.05 \\
\bottomrule
    \end{tabular}
    }
    
\end{table*}

\begin{table*}[h] 
    \scriptsize
    \tiny
    \centering
    \setlength{\tabcolsep}{2.5pt}
    \caption{\label{tab:results-sft} Full experimental results on the MMLA benchmark using supervised fine-tuning.}
    \scalebox{0.86}{
    \vspace{2mm}
    \centering
    \setlength{\belowcaptionskip}{-3.7cm}
    \begin{tabular}{l|c|>
    {\centering\arraybackslash}p{2.1cm}|>
    {\centering\arraybackslash}p{2.1cm}|>{\centering\arraybackslash}p{2.1cm}|>{\centering\arraybackslash}p{2.1cm}|>{\centering\arraybackslash}p{2.1cm}|>{\centering\arraybackslash}p{2.1cm}|>{\centering\arraybackslash}p{2.1cm}|}
        \toprule
        \addlinespace[0.1cm]
         & Datasets & \multicolumn{2}{c|}{MIntRec} & \multicolumn{2}{c|}{MIntRec2.0} & \multicolumn{2}{c|}{MELD}\\
         \addlinespace[0.1cm]
          \cline{2-8}
          \addlinespace[0.1cm]
         & Models 
         & ACC \ \hspace{0.05cm} WF1 \  WP & F1 \ \hspace{0.05cm} R \   P & ACC \ \hspace{0.05cm} WF1 \   WP & F1 \ \hspace{0.05cm} R \   P & ACC \ \hspace{0.05cm} WF1 \   WP & F1 \ \hspace{0.05cm} R \   P \\

         \midrule
         \multirow{8}{*}{\rotatebox[origin=c]{90}{LLMs}}
         & Qwen2-0.5B      & 70.11 \ 70.01 \ 71.37 & 66.97 \ 66.90 \ 68.97 & 55.83 \ 55.04 \ 55.74 & 48.08 \ 47.03 \ 51.22 & 63.95 \ 62.73 \ 62.15 & 45.04 \ 43.43 \ 47.30 \\
    & Llama-3.2-1B    & 73.26 \ 73.45 \ 73.93 & 71.19 \ 71.30 \ 71.50 & 59.37 \ 59.06 \ 60.63 & 53.49 \ 53.17 \ 56.33 & 61.57 \ 60.84 \ 60.83 & 44.91 \ 43.99 \ 47.90 \\
    & Qwen2-1.5B      & \underline{74.61} \ 74.71 \ 75.88 & \underline{72.85} \ 72.30 \ 74.96 & 58.39 \ 57.78 \ 59.06 & 52.74 \ 51.82 \ 56.25 & 64.71 \ 63.74 \ 63.57 & 47.16 \ 45.96 \ 49.89 \\
    & Llama-3.2-3B    & \textbf{77.30} \ \underline{77.17} \ \underline{77.97} & \textbf{74.40} \ 73.04 \ \textbf{77.02} & 60.45 \ 59.74 \ 60.27 & 55.28 \ 56.52 \ 56.07 & 64.41 \ 63.40 \ 63.11 & 47.59 \ 46.17 \ 50.20 \\
    & Qwen2-7B        & \underline{74.61} \ 75.10 \ 77.87 & 69.90 \ 70.26 \ 71.89 & 61.49 \ 60.91 \ 62.70 & 53.72 \ 54.29 \ 56.61 & \underline{65.71} \ \underline{65.11} \ \textbf{65.12} & \textbf{50.86} \ \textbf{49.28} \ \underline{54.90} \\
\bottomrule
    \end{tabular}
    }
    \scalebox{0.86}{
    \begin{tabular}{l|c|>
    {\centering\arraybackslash}p{2.1cm}|>
    {\centering\arraybackslash}p{2.1cm}|>{\centering\arraybackslash}p{2.1cm}|>{\centering\arraybackslash}p{2.1cm}|>{\centering\arraybackslash}p{2.1cm}|>{\centering\arraybackslash}p{2.1cm}|>{\centering\arraybackslash}p{2.1cm}|}
        \toprule
        \addlinespace[0.1cm]
         & Datasets & \multicolumn{2}{c|}{IEMOCAP} & \multicolumn{2}{c|}{MELD-DA} & \multicolumn{2}{c|}{IEMOCAP-DA}\\
                   \cline{2-8}
          \addlinespace[0.1cm]
         & Models 
         & ACC \ \hspace{0.05cm} WF1 \  WP & F1 \ \hspace{0.05cm} R \   P & ACC \ \hspace{0.05cm} WF1 \   WP & F1 \ \hspace{0.05cm} R \   P & ACC \ \hspace{0.05cm} WF1 \   WP & F1 \ \hspace{0.05cm} R \   P \\
         \midrule

         \multirow{8}{*}{\rotatebox[origin=c]{90}{LLMs}}
         & Qwen2-0.5B      & 46.61 \ 45.90 \ 47.25 & 43.61 \ 43.44 \ 46.54 & 57.31 \ 56.11 \ 56.26 & 48.98 \ 47.79 \ 53.58 & 68.21 \ 68.00 \ 68.84 & 65.01 \ 61.75 \ \textbf{75.87} \\
            & Llama-3.2-1B    & 49.08 \ 48.46 \ 49.38 & 45.99 \ 46.46 \ 47.64 & 58.36 \ 57.25 \ 57.21 & \underline{50.09} \ \underline{48.81} \ 53.16 & 69.11 \ 68.90 \ 69.39 & 63.55 \ 62.00 \ 66.01 \\
            & Qwen2-1.5B      & 49.88 \ 49.46 \ 51.22 & 47.61 \ 48.31 \ 49.21 & 57.21 \ 56.62 \ 56.66 & 49.56 \ 47.74 \ \textbf{54.17} & 67.14 \ 66.42 \ 67.78 & 58.91 \ 60.83 \ 58.58 \\
            & Llama-3.2-3B    & 50.00 \ 49.24 \ 51.04 & 47.28 \ 48.28 \ 48.89 & \underline{58.71} \ \underline{57.36} \ 56.90 & 49.70 \ 47.80 \ \underline{53.59} & 71.39 \ 70.95 \ 71.31 & 67.71 \ 68.66 \ 67.62 \\
            & Qwen2-7B        & \underline{52.10} \ \underline{50.96} \ \underline{52.20} & \underline{48.31} \ \underline{49.84} \ \underline{49.85} & 57.71 \ 56.98 \ 57.17 & 49.25 \ 46.87 \ 53.22 & 72.08 \ 72.00 \ 72.94 & 65.16 \ 65.36 \ 65.69 \\
\bottomrule
    \end{tabular}
    }
     \scalebox{0.86}{
    \begin{tabular}{l|c|>
    {\centering\arraybackslash}p{2.1cm}|>
    {\centering\arraybackslash}p{2.1cm}|>{\centering\arraybackslash}p{2.1cm}|>{\centering\arraybackslash}p{2.1cm}|>{\centering\arraybackslash}p{2.1cm}|>{\centering\arraybackslash}p{2.1cm}|>{\centering\arraybackslash}p{2.1cm}|}
        \toprule
        \addlinespace[0.1cm]
        & Datasets & \multicolumn{2}{c|}{MOSI} & \multicolumn{2}{c|}{CH-SIMS v2.0} & \multicolumn{2}{c|}{UR-FUNNY-v2}\\
                   \cline{2-8}
          \addlinespace[0.1cm]
         & Models 
         & ACC \ \hspace{0.05cm} WF1 \  WP & F1 \ \hspace{0.05cm} R \   P & ACC \ \hspace{0.05cm} WF1 \   WP & F1 \ \hspace{0.05cm} R \   P & ACC \ \hspace{0.05cm} WF1 \   WP & F1 \ \hspace{0.05cm} R \   P \\
         \midrule

         \multirow{8}{*}{\rotatebox[origin=c]{90}{LLMs}}
            & Qwen2-0.5B & 82.94 \ 82.90 \ 83.44 & 82.91 \ 83.01 \ 83.40 & 68.96 \ 68.61 \ 68.29 & 57.73 \ 57.63 \ 57.90 & 65.59 \ 65.54 \ 65.63 & 65.51 \ 65.54 \ 65.64 \\
            & Llama-3.2-1B & 82.80 \ 82.79 \ 82.95 & 82.79 \ 82.83 \ 82.92 & 64.70 \ 65.00 \ 65.70 & 54.54 \ 54.57 \ 54.88 & 69.62 \ 69.60 \ 69.72 & 69.61 \ 69.65 \ 69.69 \\
            & Qwen2-1.5B & 85.42 \ 85.42 \ 85.43 & 85.42 \ 85.41 \ 85.44 & 69.34 \ 69.49 \ 69.77 & 59.36 \ 59.39 \ 59.44 & 68.81 \ 68.81 \ 68.85 & 68.81 \ 68.83 \ 68.84 \\
            & Llama-3.2-3B & 86.44 \ 86.44 \ 86.46 & 86.44 \ 86.43 \ 86.46 & 66.25 \ 66.09 \ 65.94 & 55.19 \ 55.18 \ 55.22 & 71.43 \ 71.43 \ 71.46 & 71.43 \ 71.44 \ 71.44 \\
            & Qwen2-7B & \underline{87.32} \ \underline{87.32} \ \underline{87.32} & \underline{87.32} \ \underline{87.31} \ \underline{87.32} & \textbf{73.11} \ \textbf{71.71} \ \underline{71.36} & \textbf{61.41} \ \underline{60.62} \ \textbf{64.51} & 71.33 \ 71.33 \ 71.35 & 71.33 \ 71.34 \ 71.34 \\
\bottomrule
    \end{tabular}
    }
     \scalebox{0.86}{
    \begin{tabular}{l|c|>
    {\centering\arraybackslash}p{2.1cm}|>
    {\centering\arraybackslash}p{2.1cm}|>{\centering\arraybackslash}p{2.1cm}|>{\centering\arraybackslash}p{2.1cm}|>{\centering\arraybackslash}p{2.1cm}|>{\centering\arraybackslash}p{2.1cm}|>{\centering\arraybackslash}p{2.1cm}|}
        \toprule
        \addlinespace[0.1cm]
        & Datasets & \multicolumn{2}{c|}{MUStARD} & \multicolumn{2}{c|}{Anno-MI (client)} & \multicolumn{2}{c|}{Anno-MI (therapist)}\\
                   \cline{2-8}
          \addlinespace[0.1cm]
         & Models 
         & ACC \ \hspace{0.05cm} WF1 \  WP & F1 \ \hspace{0.05cm} R \   P & ACC \ \hspace{0.05cm} WF1 \   WP & F1 \ \hspace{0.05cm} R \   P & ACC \ \hspace{0.05cm} WF1 \   WP & F1 \ \hspace{0.05cm} R \   P \\
         \midrule

         \multirow{8}{*}{\rotatebox[origin=c]{90}{LLMs}}
            & Qwen2-0.5B & 60.87 \ 60.87 \ 60.84 & 60.84 \ 60.87 \ 60.91 & 62.98 \ \textbf{60.96} \ 61.87 & \textbf{52.69} \ \textbf{51.37} \ 56.95 & 73.33 \ 73.33 \ 74.51 & 71.34 \ 71.47 \ 72.47 \\
            & Llama-3.2-1B & \underline{65.94} \ \underline{65.94} \ \underline{65.53} & \underline{65.53} \ \underline{65.94} \ \underline{66.73} & 61.29 \ 55.87 \ 58.20 & 42.10 \ 42.98 \ 54.74 & 74.47 \ 74.50 \ 75.17 & 72.77 \ 73.04 \ 73.13 \\
            & Qwen2-1.5B & 63.04 \ 63.04 \ 63.03 & 63.03 \ 63.04 \ 63.07 & 62.89 \ 58.90 \ 61.09 & 48.33 \ 46.96 \ 58.27 & 74.30 \ 74.44 \ 75.28 & 72.91 \ 73.50 \ 72.96 \\
            & Llama-3.2-3B & 63.04 \ 62.60 \ 63.69 & 62.6 \ 63.04 \ 63.69 & \underline{64.75} \ 60.16 \ \textbf{64.96} & 50.28 \ 48.38 \ \textbf{65.06} & 73.33 \ 73.71 \ 75.93 & 72.48 \ 73.46 \ 73.04 \\
            & Qwen2-7B & 63.04 \ 63.04 \ 63.05 & 63.04 \ 63.04 \ 63.05 & 64.57 \ \underline{60.88} \ 63.18 & \underline{50.56} \ \underline{48.87} \ 61.41 & 75.53 \ 75.80 \ 77.49 & 74.67 \ 75.39 \ \underline{75.16} \\
\bottomrule
    \end{tabular}
    }
    
\end{table*}

\begin{table*}[h] 
    \scriptsize
    \tiny
    \centering
    \setlength{\tabcolsep}{2.5pt}
    \caption{\label{tab:results-it} Full experimental results on the MMLA benchmark using instruction tuning.}
    \scalebox{1.2}{
    \vspace{2mm}
    \centering
    \setlength{\belowcaptionskip}{-3.7cm}
    \begin{tabular}{c|c|  % 修改第一列为居中对齐
    >{\centering\arraybackslash}p{2.1cm}|
    >{\centering\arraybackslash}p{2.1cm}|
    >{\centering\arraybackslash}p{2.1cm}|
    >{\centering\arraybackslash}p{2.1cm}|
    >{\centering\arraybackslash}p{2.1cm}|}
        \toprule
        \addlinespace[0.1cm]
         & Datasets & \multicolumn{2}{c|}{MIntRec} & \multicolumn{2}{c|}{MIntRec2.0}\\
         \addlinespace[0.1cm]
          \cline{2-6}  % 调整cline范围匹配列数
          \addlinespace[0.1cm]
        
         & Models 
         & ACC \ \ WF1 \ \ WP 
         & F1 \ \ R \ \ P 
         & ACC \ \ WF1 \ \ WP 
         & F1 \ \ R \ \ P \\

         \midrule
         \multirow{3}{*}{\rotatebox[origin=c]{90}{LLMs}}

            & Qwen2-7B & \underline{77.53} \ \underline{77.78} \ \underline{79.08} & \underline{74.66} \ \underline{75.42} \ 75.57 & \textbf{61.49} \ \underline{61.07} \ \underline{62.56} & \textbf{57.48} \ \textbf{56.55} \ \textbf{61.02}  \\
            
            & Llama-3-8B & \textbf{78.20} \ \textbf{78.14} \ \textbf{79.94} & \textbf{76.62} \ \textbf{76.46} \ \textbf{79.43} & \underline{61.44} \ \textbf{61.33} \ \textbf{63.12} & \underline{57.05} \ \underline{56.28} \ \underline{60.05} \\
            & Internlm-2.5-7B & 75.96 \ 76.06 \ 78.46 & 73.11 \ 71.81 \ \underline{76.71} & 60.06 \ 59.35 \ 62.27 & 54.13 \ 52.69 \ 58.83 \\
\bottomrule
    \end{tabular}
    }

    \scalebox{1.2}{
    \vspace{2mm}
    \centering
    \setlength{\belowcaptionskip}{-3.7cm}
    \begin{tabular}{c|c|  % 修改第一列为居中对齐
    >{\centering\arraybackslash}p{2.1cm}|
    >{\centering\arraybackslash}p{2.1cm}|
    >{\centering\arraybackslash}p{2.1cm}|
    >{\centering\arraybackslash}p{2.1cm}|
    >{\centering\arraybackslash}p{2.1cm}|}
        \toprule
        \addlinespace[0.1cm]
         & Datasets & \multicolumn{2}{c|}{MELD} & \multicolumn{2}{c|}{IEMOCAP}\\
         \addlinespace[0.1cm]
          \cline{2-6}  % 调整cline范围匹配列数
          \addlinespace[0.1cm]
        
         & Models 
         & ACC \ \ WF1 \ \ WP 
         & F1 \ \ R \ \ P 
         & ACC \ \ WF1 \ \ WP 
         & F1 \ \ R \ \ P \\

         \midrule
         \multirow{3}{*}{\rotatebox[origin=c]{90}{LLMs}}

            % & Qwen2-7B & 64.64 \ 64.09 \ 65.15 & 48.83 \ 48.05 \ 52.81 & 48.09 \ 47.00 \ 52.49 & 45.53 \ 47.99 \ 49.46 \\
            % & Llama-3-8B & 66.02 \ 64.79 \ 65.19 & 49.86 \ 48.77 \ 53.82 & 52.34 \ 51.28 \ 56.88 & 51.04 \ 53.07 \ 54.89 \\
            % & Internlm-2.5-7B & 62.84 \ 62.93 \ 63.71 & 49.76 \ 49.90 \ 51.54 & 50.80 \ 50.21 \ 54.10 & 49.56 \ 50.75 \ 52.31 \\

        & Qwen2-7B 
        & \underline{64.64} \ \underline{64.09} \ \underline{65.15} 
        & 48.83 \ 48.05 \ \underline{52.81} 
        & 48.09 \ 47.00 \ 52.49
        & 45.53 \ 47.99 \ 49.46 \\
        
        & Llama-3-8B 
        & \textbf{66.02} \ \textbf{64.79} \ \textbf{65.19} 
        & \textbf{49.86} \ \underline{48.77} \ \textbf{53.82} 
        & \textbf{52.34} \ \textbf{51.28} \ \textbf{56.88} 
        & \textbf{51.04} \ \textbf{53.07} \ \textbf{54.89} \\
        
        & Internlm-2.5-7B 
        & 62.84 \ 62.93 \ 63.71 
        & \underline{49.76} \ \textbf{49.90} \ 51.54 
        & \underline{50.80} \ \underline{50.21} \ \underline{54.10} 
        & \underline{49.56} \ \underline{50.75} \ \underline{52.31} \\
\bottomrule
    \end{tabular}
    }

    \scalebox{1.2}{
    \vspace{2mm}
    \centering
    \setlength{\belowcaptionskip}{-3.7cm}
    \begin{tabular}{c|c|  % 修改第一列为居中对齐
    >{\centering\arraybackslash}p{2.1cm}|
    >{\centering\arraybackslash}p{2.1cm}|
    >{\centering\arraybackslash}p{2.1cm}|
    >{\centering\arraybackslash}p{2.1cm}|
    >{\centering\arraybackslash}p{2.1cm}|}
        \toprule
        \addlinespace[0.1cm]
         & Datasets & \multicolumn{2}{c|}{MELD-DA} & \multicolumn{2}{c|}{IEMOCAP-DA}\\
         \addlinespace[0.1cm]
          \cline{2-6}  % 调整cline范围匹配列数
          \addlinespace[0.1cm]
        
         & Models 
         & ACC \ \ WF1 \ \ WP 
         & F1 \ \ R \ \ P 
         & ACC \ \ WF1 \ \ WP 
         & F1 \ \ R \ \ P \\

         \midrule
         \multirow{3}{*}{\rotatebox[origin=c]{90}{LLMs}}

            % & Qwen2-7B & 58.71 \ 57.46 \ 57.54 & 48.81 \ 48.48 \ 52.23 & 65.50 \ 64.91 \ 66.69 & 60.28 \ 63.67 \ 63.60 \\
            % & Llama-3-8B & 58.91 \ 57.44 \ 57.26 & 47.53 \ 45.89 \ 52.29 & 69.06 \ 68.36 \ 69.38 & 62.65 \ 64.99 \ 64.40 \\
            % & Internlm-2.5-7B & 57.51 \ 55.80 \ 54.94 & 45.16 \ 44.79 \ 47.29 & 67.57 \ 66.91 \ 67.60 & 59.69 \ 61.16 \ 61.67 \\

            & Qwen2-7B 
            & \underline{58.71} \ \textbf{57.46} \ \textbf{57.54} 
            & \textbf{48.81} \ \textbf{48.48} \ \underline{52.23} 
            & 65.50 \ 64.91 \ 66.69
            & \underline{60.28} \ \underline{63.67} \ \underline{63.60} \\
            
            & Llama-3-8B 
            & \textbf{58.91} \ \underline{57.44} \ \underline{57.26} 
            & \underline{47.53} \ \underline{45.89} \ \textbf{52.29} 
            & \textbf{69.06} \ \textbf{68.36} \ \textbf{69.38} 
            & \textbf{62.65} \ \textbf{64.99} \ \textbf{64.40} \\
            
            & Internlm-2.5-7B 
            & 57.51 \ 55.80 \ 54.94 
            & 45.16 \ 44.79 \ 47.29 
            & \underline{67.57} \ \underline{66.91} \ \underline{67.60} 
            & 59.69 \ 61.16 \ 61.67 \\
\bottomrule
    \end{tabular}
    }

    \scalebox{1.2}{
    \vspace{2mm}
    \centering
    \setlength{\belowcaptionskip}{-3.7cm}
    \begin{tabular}{c|c|  % 修改第一列为居中对齐
    >{\centering\arraybackslash}p{2.1cm}|
    >{\centering\arraybackslash}p{2.1cm}|
    >{\centering\arraybackslash}p{2.1cm}|
    >{\centering\arraybackslash}p{2.1cm}|
    >{\centering\arraybackslash}p{2.1cm}|}
        \toprule
        \addlinespace[0.1cm]
         & Datasets & \multicolumn{2}{c|}{MOSI} & \multicolumn{2}{c|}{CH-SIMS v2.0}\\
         \addlinespace[0.1cm]
          \cline{2-6}  % 调整cline范围匹配列数
          \addlinespace[0.1cm]

         & Models 
         & ACC \ \ WF1 \ \ WP 
         & F1 \ \ R \ \ P 
         & ACC \ \ WF1 \ \ WP 
         & F1 \ \ R \ \ P \\

         \midrule
         \multirow{3}{*}{\rotatebox[origin=c]{90}{LLMs}}

            & Qwen2-7B & \underline{87.17} \ \underline{87.17} \ \underline{87.20} & \underline{87.17} \ \underline{87.18} \ \underline{87.18} & \textbf{70.50} \ 69.59 \ 68.88 & 58.09 \ 57.97 \ 58.69 \\
            & Llama-3-8B & \textbf{87.32} \ \textbf{87.31} \ \textbf{87.35} & \textbf{87.31} \ \textbf{87.30} \ \textbf{87.36} & 69.54 \ \underline{70.04} \ \underline{70.62} & \textbf{60.50} \ \textbf{60.86} \ \textbf{60.32} \\
            & Internlm-2.5-7B & 86.01 \ 86.00 \ 86.02 & 86.00 \ 85.99 \ 86.03 & \underline{69.63} \ \textbf{70.06} \ \textbf{70.69} & \underline{59.45} \ \underline{59.57} \ \underline{59.58} \\
            
        \midrule
        \multirow{3}{*}{\rotatebox[origin=c]{90}{MLLMs}} 
 
            & Qwen2-VL-7B & 79.45 \ 83.78 \ \textbf{89.74} & 55.82 \ 52.89 \ 59.86 & \underline{75.85} \ \underline{73.05} \ 73.07 & \underline{59.54} \ 59.82 \ \underline{65.70} \\
            & MiniCPM-V-2.6-8B & \underline{86.15} \ \underline{86.08} \ \underline{87.06} & \underline{86.09} \ \underline{86.24} \ \textbf{86.99} & \textbf{79.15} \ \textbf{74.99} \ \textbf{79.88} & 58.95 \ \underline{61.00} \ \textbf{81.29} \\
            & Qwen2-VL-72B & 81.63 \ 81.31 \ 84.25 & 81.34 \ 81.79 \ 84.14 & 72.65 \ 68.26 \ 67.52 & 51.56 \ 55.53 \ 50.45 \\
\bottomrule
    \end{tabular}
    }

    \scalebox{1.2}{
    \vspace{2mm}
    \centering
    \setlength{\belowcaptionskip}{-3.7cm}
    \begin{tabular}{c|c|  % 修改第一列为居中对齐
    >{\centering\arraybackslash}p{2.1cm}|
    >{\centering\arraybackslash}p{2.1cm}|
    >{\centering\arraybackslash}p{2.1cm}|
    >{\centering\arraybackslash}p{2.1cm}|
    >{\centering\arraybackslash}p{2.1cm}|}
        \toprule
        \addlinespace[0.1cm]
         & Datasets & \multicolumn{2}{c|}{UR-FUNNY-v2} & \multicolumn{2}{c|}{MUStARD}\\
         \addlinespace[0.1cm]
          \cline{2-6}  % 调整cline范围匹配列数
          \addlinespace[0.1cm]
        
         & Models 
         & ACC \ \ WF1 \ \ WP 
         & F1 \ \ R \ \ P 
         & ACC \ \ WF1 \ \ WP 
         & F1 \ \ R \ \ P \\

         \midrule
         \multirow{3}{*}{\rotatebox[origin=c]{90}{LLMs}}

            & Qwen2-7B & \underline{70.22} \ \underline{70.22} \ \underline{70.22} & \underline{70.21} \ \underline{70.21} \ \underline{70.22} & \underline{63.04} \ \underline{61.34} \ \underline{65.84} & \underline{61.34} \ \underline{63.04} \ \underline{65.84} \\
            & Llama-3-8B & \textbf{72.13} \ \textbf{72.12} \ \textbf{72.15} & \textbf{72.10} \ \textbf{72.10} \ \textbf{72.16} & 57.97 \ 55.25 \ 60.53 & 55.25 \ 57.97 \ 60.53 \\
            & Internlm-2.5-7B & 69.11 \ 68.99 \ 69.57 & 69.01 \ 69.21 \ 69.53 & \textbf{65.94} \ \textbf{64.59} \ \textbf{68.82} & \textbf{64.59} \ \textbf{65.94} \ \textbf{68.82} \\
            
        \midrule
        \multirow{3}{*}{\rotatebox[origin=c]{90}{MLLMs}} 
 
            & Qwen2-VL-7B & 73.34 \ 73.28 \ 73.46 & 73.26 \ 73.28 \ 73.48 & 65.22 \ 65.21 \ 65.23 & 65.21 \ 65.22 \ 65.23 \\
            & MiniCPM-V-2.6-8B & 74.14 \ \underline{74.15} \ 74.15 & \underline{74.14} \ 74.15 \ 74.14 & 60.87 \ 60.84 \ 60.91 & 60.84 \ 60.87 \ 60.91 \\
            & Qwen2-VL-72B & \textbf{76.96} \ \textbf{76.35} \ \textbf{80.48} & \textbf{76.40} \ \textbf{77.19} \ \textbf{80.33} & \underline{70.29} \ \underline{68.49} \ \textbf{76.31} & \underline{68.49} \ \underline{70.29} \ \textbf{76.31} \\
\bottomrule
    \end{tabular}
    }

    \scalebox{1.2}{
    \vspace{2mm}
    \centering
    \setlength{\belowcaptionskip}{-3.7cm}
    \begin{tabular}{c|c|  % 修改第一列为居中对齐
    >{\centering\arraybackslash}p{2.1cm}|
    >{\centering\arraybackslash}p{2.1cm}|
    >{\centering\arraybackslash}p{2.1cm}|
    >{\centering\arraybackslash}p{2.1cm}|
    >{\centering\arraybackslash}p{2.1cm}|}
        \toprule
        \addlinespace[0.1cm]
         & Datasets & \multicolumn{2}{c|}{Anno-MI (client)} & \multicolumn{2}{c|}{Anno-MI (therapist)}\\
         \addlinespace[0.1cm]
          \cline{2-6}  % 调整cline范围匹配列数
          \addlinespace[0.1cm]
        
         & Models 
         & ACC \ \ WF1 \ \ WP 
         & F1 \ \ R \ \ P 
         & ACC \ \ WF1 \ \ WP 
         & F1 \ \ R \ \ P \\

         \midrule
         \multirow{3}{*}{\rotatebox[origin=c]{90}{LLMs}}

            & Qwen2-7B & \underline{64.04} \ \underline{61.00} \ 62.87 & \underline{52.00} \ \textbf{50.43} \ 58.88 & \underline{72.28} \ \underline{71.61} \ \underline{74.19} & 69.37 \ \underline{69.89} \ \underline{72.78} \\
            & Llama-3-8B & \textbf{66.16} \ \textbf{61.34} \ \textbf{68.87} & \textbf{52.60} \ \underline{50.38} \ \textbf{69.72} & \textbf{75.26} \ \textbf{75.40} \ \textbf{76.26} & \textbf{74.21} \ \textbf{75.00} \ \textbf{74.06} \\
            & Internlm-2.5-7B & 62.18 \ 54.40 \ \underline{64.20} & 40.80 \ 41.92 \ \underline{67.05} & 71.23 \ 71.42 \ 73.42 & \underline{69.54} \ 69.25 \ 71.73 \\
            
        \midrule
        \multirow{3}{*}{\rotatebox[origin=c]{90}{MLLMs}} 
 
            & Qwen2-VL-7B & 64.54 \ \underline{60.34} \ 62.21 & \underline{48.41} \ \underline{47.60} \ 58.61 & 76.44 \ 76.65 \ 77.58 & 75.14 \ 75.26 \ 75.68 \\
            & MiniCPM-V-2.6-8B & 64.48 \ 57.16 \ \underline{71.31} & 42.60 \ 43.75 \ \textbf{78.95} & 75.18 \ 75.27 \ \underline{77.83} & 74.00 \ 74.96 \ 75.30 \\
            & Qwen2-VL-72B & \underline{64.89} \ 57.99 \ \textbf{72.12} & 46.01 \ 45.35 \ \underline{76.80} & \underline{76.93} \ \underline{76.82} \ 77.73 & \underline{75.46} \ \underline{75.88} \ \underline{76.11} \\
\bottomrule
    \end{tabular}
    }
\end{table*}

\begin{table*}[!t]
\small
\caption{\label{tab:param-1} Primary hyperparameters for SFT and IT on the MIntRec, MIntRec2.0, and MELD datasets.}
\centering
\setlength{\belowcaptionskip}{-3.7cm}
\scalebox{0.85}{
\begin{tabular}{c|ccccccccc}
\toprule
Datasets & Methods & Models & Learning rate & Warmup ratio & Training Batch size & Epochs & Rank & $\alpha$ \\
\midrule
     \multirow{19}{*}{\rotatebox[origin=c]{90}{MIntRec}}  
     &  \multirow{11}{*}{SFT} 
     & Llama3-8B & 0.0001 & 0.1 & 2 & 8 & 8 & 16 \\
     & & Qwen2-7B & 0.0001 & 0.1 & 2 & 8 & 8 & 16 \\
     & & InternLM2.5-7B & 0.0001 & 0.1 & 2 & 8 & 8 & 16 \\
     & & VideoLLaMA2-7B & 0.00002 & 0.5 & 1 & 10 & 128 & 256 \\
     & & Qwen2-VL-7B  & 0.00011 & 0.1 & 8 & 60 & 8 & 16 \\
     & & LLaVA-Video-7B & 0.0002 & 0.3 & 4 & 10 & 128 & 256 \\
     & & LLaVA-OV-7B & 0.0001 & 0.1 & 6 & 15 & 128 & 256 \\
\bottomrule
\end{tabular}
}
\end{table*}

\begin{table*}[!t]
\small
\caption{\label{tab:param-2} Primary hyperparameters for SFT and IT on the IEMOCAP, MELD-DA, and IEMOCAP-DA datasets.}
\centering
\setlength{\belowcaptionskip}{-3.7cm}
\scalebox{0.85}{
\begin{tabular}{c|ccccccccc}
\toprule
Datasets & Methods & Models & Learning rate & Warmup ratio & Training Batch size & Epochs & Rank & $\alpha$ \\
\midrule
     \multirow{19}{*}{\rotatebox[origin=c]{90}{IEMOCAP}}  
    &  \multirow{11}{*}{SFT}  
    & Llama3-8B & 0.0001 & 0.1 & 2 & 8 & 8 & 16 \\
     & & Qwen2-7B & 0.0001 & 0.1 & 2 & 8 & 8 & 16 \\
     & & InternLM2.5-7B & 0.0001 & 0.1 & 2 & 8 & 8 & 16 \\
    & & VideoLLaMA2-7B & 0.00002 & 0.5 & 1 & 10 & 128 & 256 \\
    & & Qwen2-VL-7B & 0.0001 & 0.2 & 2 & 15 & 8 & 16 \\
    & & LLaVA-Video-7B & 0.0002 & 0.2 & 6 & 8 & 128 & 256 \\
    & & LLaVA-OV-7B & 0.0001 & 0.1 & 6 & 10 & 128 & 256 \\
\bottomrule
\end{tabular}
}
\end{table*}

\begin{table*}[!t]
\small
\caption{\label{tab:param-3} Primary hyperparameters for SFT and IT on the MOSI, CH-SIMS v2.0, and UR-FUNNY-v2 datasets.}
\centering
\setlength{\belowcaptionskip}{-3.7cm}
\scalebox{0.85}{
\begin{tabular}{c|ccccccccc}
\toprule
Datasets & Methods & Models & Learning rate & Warmup ratio & Training Batch size & Epochs & Rank & $\alpha$ \\
\midrule
     \multirow{19}{*}{\rotatebox[origin=c]{90}{MOSI}}  
    &  \multirow{11}{*}{SFT}  
    & Llama3-8B & 0.0001 & 0.1  & 2 & 8 & 8 & 16  \\
     & & Qwen2-7B & 0.0001 & 0.1  & 2 & 8 & 8 & 16  \\
     & & InternLM2.5-7B & 0.0001 & 0.1  & 2 & 8 & 8 & 16  \\
    & & VideoLLaMA2-7B & 0.00002 & 0.5 & 1 & 10 & 128 & 256 \\
    & & Qwen2-VL-7B & 0.0001 & 0.1 & 8 & 20 & 8 & 16 \\
    & & LLaVA-Video-7B & 0.0002 & 0.3 & 6 & 5 & 128 & 256 \\
    & & LLaVA-OV-7B & 0.0001 & 0.1 & 6 & 20 & 128 & 256 \\
\bottomrule
\end{tabular}
}
\end{table*}

\begin{table*}[!t]
\small
\caption{\label{tab:param-4} Primary hyperparameters for SFT and IT on the MUStARD, Anno-MI (client), and Anno-MI (therapist) datasets.}
\centering
\setlength{\belowcaptionskip}{-3.7cm}
\scalebox{0.85}{
\begin{tabular}{c|ccccccccc}
\toprule
Datasets & Methods & Models & Learning rate & Warmup ratio & Training Batch size & Epochs & Rank & $\alpha$ \\
\midrule
     \multirow{19}{*}{\rotatebox[origin=c]{90}{MUStARD}}  
    &  \multirow{11}{*}{SFT}  
    & Llama3-8B & 0.00005 & 0.1 & 1 & 8 & 16 & 32 \\
     & & Qwen2-7B & 0.00005 & 0.1 & 1 & 8 & 16 & 32  \\
     & & InternLM2.5-7B & 0.00005 & 0.1 & 1 & 8 & 16 & 32 \\
    & & VideoLLaMA2-7B & 0.00002 & 0.5 & 1 & 10 & 128 & 256 \\
    & & Qwen2-VL-7B & 0.001 & 0.2 & 2 & 10 & 8 & 16 \\
    & & LLaVA-Video-7B & 0.0002 & 0.4 & 2 & 10 & 128 & 256 \\
    & & LLaVA-OV-7B & 0.0001 & 0.1 & 6 & 15 & 128 & 256 \\
\bottomrule
\end{tabular}
}
\end{table*}